\documentclass[a4paper,fleqn]{cas-dc}

\usepackage[authoryear]{natbib}
\usepackage{microtype}
\usepackage{graphicx}
\usepackage{booktabs} 
\usepackage{enumitem}
\usepackage{multirow}
\usepackage{xcolor}

\usepackage{amsmath}
\usepackage{amssymb}
\usepackage{mathtools}
\usepackage{amsthm}
\usepackage{subcaption}
\usepackage{amsfonts}

\def\tsc#1{\csdef{#1}{\textsc{\lowercase{#1}}\xspace}}
\tsc{WGM}
\tsc{QE}

\newcommand{\model}{GAIR}

\newcommand{\rxi}{z_{i}^{(q)}}
\newcommand{\rxj}{z_{j}^{(q)}}
\newcommand{\inrfun}{\Omega}
\newcommand{\rsArea}{A}

\newcommand{\simf}{\mathcal{D}}

\newcommand{\rone}[1]{\textcolor{black}{{#1}}}

\begin{document}
\let\WriteBookmarks\relax

\shorttitle{}    

\shortauthors{}  

\title [mode = title]{\model: Location-Aware Self-Supervised Contrastive Pre-Training with Geo-Aligned Implicit Representations}



%








\author[1]{Zeping Liu}
[orcid=0000-0003-2898-0023]
\ead{zeping.liu@utexas.edu}

\author[2,1]{Ni Lao}
[orcid=0000-0002-4034-7784]
\ead{nlao@google.com}

\author[3]{Zhangyu Wang}
[orcid=0009-0004-4728-4458]
\ead{zhangyu.wang@maine.edu}

\author[4]{Junfeng Jiao}
[orcid=0000-0002-7272-8805]
\ead{jjiao@austin.utexas.edu}

\author[1]{Gengchen Mai}
[orcid=0000-0002-7818-7309]
\cormark[1] 
\ead{gengchen.mai@utexas.edu}
\cortext[cor1]{Corresponding author}

\address[1]{SEAI Lab, Department of Geography and the Environment, The University of Texas at Austin, Austin, TX, USA 78712}
\address[2]{Google LLC, Mountain View, CA, USA 94043}
\address[3]{SIT Lab, School of Computing and Information Science, The University of Maine, Orono, ME, USA 04101}
\address[4]{Urban Information Lab, School of Architecture, The University of Texas at Austin, Austin, TX, USA 78712}

\begin{abstract}
\textbf{Vision Transformer (ViT)} has been widely used in computer vision tasks with excellent results by providing representations for a whole image or image patches. However, ViT lacks \textit{detailed localized image representations} 
at arbitrary positions 
when applied to geospatial tasks that involve multiple geospatial data modalities, such as overhead remote sensing (RS) data, ground-level imagery, and geospatial vector data. 
Here
high-resolution localized representations 
are vital for modeling geospatial relationships and alignments across modalities.
We proposed to solve this representation problem
with an implicit neural representation (\textbf{INR}) module extending ViT with Neural Implicit Local Interpolation, which
produces a continuous RS image representation covering
arbitrary location in the RS image.
Based on the INR module, we introduce \textbf{\model{}}, a 
novel location-aware self-supervised learning (SSL) objective
integrating overhead RS data, street view (SV) imagery, and their geolocation metadata. 
\model{} utilizes three factorized neural encoders 
to project different modalities
into the embedding space, and the INR module is used to further align these representations geographically, which
are trained with contrastive learning objectives from unlabeled data. 
We evaluate \model{} across 9 geospatial tasks and 22 datasets spanning RS image-based, SV image-based, and location embedding-based benchmarks. Experimental results demonstrate that \model{} outperforms state-of-the-art geo-foundation models (GeoFM) and 
alternative SSL training objectives (e.g., MoCo V3 and MAE) that do not use fine-grained geo-aligned spatial representations. 
Our results highlight the 
effectiveness of \model{} in 
\textbf{learning generalizable geospatial representations across tasks, spatial scales, and temporal contexts.} \rone{The project code is available at \url{https://github.com/zpl99/GAIR}}.

\end{abstract}


\begin{highlights}
\item \model{} bridges the scale gap between satellite and street-view imagery via a geo-aligned contrastive learning framework.
\item A novel Neural Implicit Local Interpolation (NILI) module enables continuous, coordinate-level alignment across heterogeneous modalities.
\item Pretrained on the new Streetscapes1M dataset, GAIR outperforms existing GeoFMs on 9 downstream tasks.
\end{highlights}



\maketitle

\section{Introduction}
\label{intro}
\begin{figure}
    \centering
    \includegraphics[width=1\linewidth]{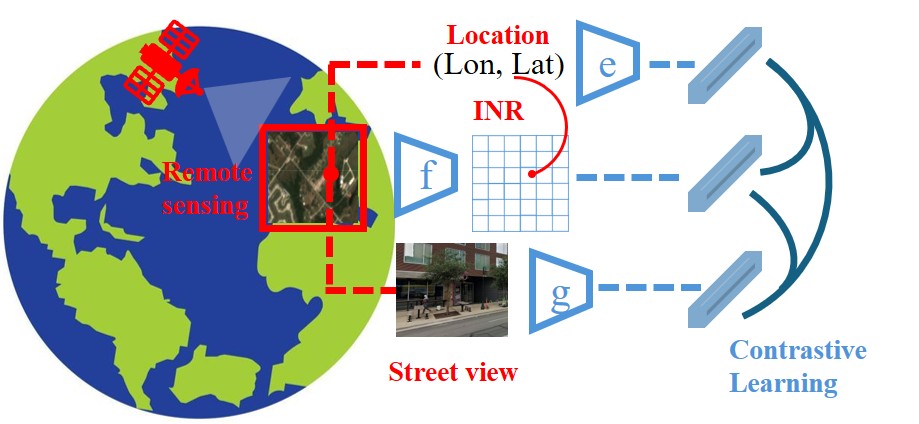}
    \caption{Overview of the \textbf{\model{}} architecture. 
    The model encodes three modalities: street view image \( s_i \), geolocation \( x_i \), and remote sensing image \( r_i \). 
    ViT-based encoders extract features \( g(s_i) \) and \( f(r_i) \), while a location encoder maps \( x_i \) to \( e(x_i) \). 
    An INR module refines \( f(r_i) \) into a geo-aligned embedding \( z_i^{(q)} \), which is used for contrastive learning with \( g(s_i) \) and \( e(x_i) \).}
    \label{fig:fig1}
\end{figure}

In the geospatial domain, 
there is a vast amount of \textbf{unlabeled geospatial datasets} such as satellite images, street view images, and user-generated geo-tagged data (e.g., Flickr images, geo-tagged tweets, iNaturalist species images, 
etc). 
In contrast, \textbf{labeled geospatial data} is typically scarce and highly imbalanced in terms of spatial, temporal, and class coverage \citep{mai2023csp,klemmer2023satclip} due to the high cost of data annotation and the specialized domain expertise required. 
This scarcity of labeled data significantly limits the usage of these multimodal geospatial data in critical geospatial applications such as economic development prediction \citep{jean2016combining}, species distribution modeling \citep{mai2023sphere2vec,cole2023spatial}, crop yield estimation \citep{azzari2017towards,you2017deep}, urban dynamics monitoring \citep{cai2020traffic}, geographic question answering \citep{mai2020se,yu2025spatial}, and climate extreme event detection \citep{ham2019deep}. Furthermore, training AI models on these limited labeled datasets constrains their generalizability across space \citep{li2022improving,li2023rethink,wu2024torchspatial}, time, and task \citep{cong2022satmae,mai2025towards}. 
%
Meanwhile, inspired by the recent advancements of language and vision foundation models, geo-foundation models (GeoFMs) \citep{mai2024opportunities,hsu2024geospatial,janowicz2025geofm} are developed as task-agnostic pre-trained models to tackle the issue of limited supervision information. 
However, \textit{existing GeoFMs heavily rely on overhead remote sensing (RS) tasks }\citep{cong2022satmae,guo2024skysense,hong2024spectralgpt}, e.g., RS foundation models (RSFMs).
Although multiple multimodal GeoFMs 
\citep{fuller2024croma, astruc2024omnisat} have been developed, they primarily focus on integrating different modalities of overhead RS data (e.g., optical, synthetic aperture radar (SAR), etc.) and/or language modality \citep{kuckreja2024geochat,zhang2024earthgpt} while overlooking other important geospatial data modalities such as ground-level images and geospatial vector data. 
This practice 
significantly limits the spatial reasoning capabilities of these models and hampers their generalizability across different tasks, regions, spatial scales, and temporal contexts. 
Furthermore, existing GeoFMs mainly rely on conventional language and vision foundation model objectives like temporal or spatial augmented embeddings \citep{ayush2021geography,stojnic2021self, guo2024skysense} or image reconstruction \citep{cong2022satmae,sun2022ringmo,jain2022multimodal}, without explicitly considering the precise spatial alignment across different data modalities. This lack of explicit geo-alignment mechanism leads to suboptimal results for geospatial applications that require fine-grained spatial understanding.

A key challenge for the current RS-based GeoFMs \citep{guo2024skysense,hong2024spectralgpt} and geo-aware SSL approaches \citep{ayush2021geography,cong2022satmae,klemmer2023satclip,mai2023csp} lies in 
aligning geospatial data across diverse spatial scales and modeling spatial relations explicitly. 
For instance, overhead satellite imagery provides a broad spatial contextual view, while street view (SV) images capture fine-grained, ground-level details. This creates an intrinsic \textbf{one-to-many} spatial relationship, where a single 256$\times$256 overhead Sentinel-2 RS image can correspond to hundreds or even thousands of SV images at different locations. To bridge this gap, a common practice to align both modalities is to construct a set of geographically co-located RS-SV image pairs, which has been widely used in cross-view image geolocalization \citep{workman2015wide,wang2023fine,li2025cross} and cross-view image synthesis \citep{regmi2018cross,wu2022cross,ye2025leveraging} tasks. However, this strategy is fundamentally limited by resolution constraints, it typically requires very high-resolution RS images (e.g., 15 cm - 1 m spatial resolution) \citep{workman2015wide,li2025cross} to match SV images' spatial scale. In easily accessible medium-resolution RS data (e.g., Sentinel-2), the viewshed of an SV image might collapse to a sub-pixel scale of RS data, making it fundamentally challenging to construct effective co-located image pairs without losing fine-grained spatial correspondences. Furthermore, such paring practice treats RS-SV image pairs as independent training samples and neglects the geospatial relation among them, while relying on neural networks to infer such structure implicitly. This practice will lead to poor geospatial reasoning ability and reduced geographic generalizability.

To tackle the above challenges, 
in this paper, 
inspired by resolution-agnostic image super-resolution methods such as LIIF \citep{chen2021learning},
we propose Geo-Aligned Implicit Representations, namely \model{}, a novel location-aware self-supervised pretraining framework. 
To bridge the spatial scale difference between RS and SV images, 
\model{} leverages neural implicit functions (NIFs) \citep{sitzmann2020implicit,chen2021learning,gao2023implicit} to learn a continuous image representation of an RS 
image and extract a localized RS neural representation that is geographically colocated with an SV image. Due to the resolution-agnostic nature of NIFs, these localized RS representations can dynamically adapt to the spatial scale of their corresponding SV images, eliminating the need to explicitly construct co-located scale-aligned RS–SV image pairs as in prior work \citep{li2025cross,ye2025leveraging}. 
Furthermore, to explicitly model the spatial relationships among different data modalities and data samples, 
we project the SV 
image and its geolocation into an SV image embedding and a location embedding \citep{mai2020multi,mai2023csp,klemmer2023satclip,wu2024torchspatial} with respective neural encoders. We then formulate self-supervised learning objectives by performing multi-object contrastive learning on these three geo-aligned embeddings, enabling \model{} to learn spatially grounded and scale-consistent representations. 
The overall architecture of \model{} is illustrated in Figure \ref{fig:fig1}. 


The contributions of this paper are as follows:
\begin{itemize}
\item We propose \model{}, a geo-aligned multimodal SSL framework capable of handling diverse geospatial modalities with diverse spatial scales, including remote sensing imagery, street view imagery, and geo-locations. By explicitly aligning these modalities geographically 
\model{} can perform a more comprehensive geospatial understanding and reasoning. 
\item  The design of \model{} involves three key technical components: a) Factorized multimodal encoder design allows the model to retain modality-specific information while effectively learning cross-modal relationships. 
b) Neural implicit functions learn continuous representations on RS images and extract fine-grained localized RS embeddings that are geographically co-located with ground-level images to bridge their spatial scale difference. 
c) 
Contrastive learning is utilized on these geo-aligned neural embeddings from three distinct modalities in order to learn generalizable geospatial representations across different tasks. 
\item We pre-train \model{} on a globally sampled dataset named \textbf{Streetscapes1M} with 1 million sampled tuples. The pre-trained \model{} is adapted to a wide range of geospatial tasks via few-shot learning or fully fine-tuning. Experimental results show that \model{} outperforms all baselines and achieves state-of-the-art (SOTA) performance across all 9 geospatial tasks and 22 datasets, including street view imagery tasks, remote sensing imagery tasks, and location embedding tasks. 
\item Further analysis shows that our geo-alignment SSL strategy (i.e., \model{}) is crucial to achieving superior performance on single-modal tasks, and multimodal fusion can further add 4-20\% performance gain. While Streetscapes1M is biased towards urban areas, by further pretraining \model{} on an urban-rural balanced dataset, the geographic bias of \model{} can be significantly reduced without sacrificing performance. Further qualitative and quantitative analysis show that \model{} can capture spatial relations across modalities, which is critical for diverse tasks such as image geolocalization.
\end{itemize}

\section{Related Work}\label{related}

\subsection{Geo-Foundation Models}
Many recent Geo-Foundation Models (GeoFMs) draw inspiration from Vision Foundation Models \citep{deng2009imagenet, chen2020simple, caron2021emerging, grill2020bootstrap, liu2021swin, chen2021mocov3, he2022masked, oquab2023dinov2} and Vision-Language Models \citep{radford2021learning, zhang2021vinvl, wang2023image}. Unlike traditional vision datasets, geospatial data inherently integrates spatial and temporal information, requiring specialized model architectures. Current GeoFMs can be roughly categorized into (1) Remote Sensing Foundation Models (RSFMs), (2) Weather and Climate Foundation Models, and (3) geospatial vision-language foundation models (GeoVLFMs). RSFMs leverage contrastive learning or masked image modeling (MIM) to learn task-agnostic neural representations. GASSL \citep{ayush2021geography} and SeCo \citep{manas2021seasonal} use temporal augmentations with a MoCo v2 \citep{he2020momentum} self-supervised learning (SSL) objective, while Dino-MC \citep{wanyan2023dino} and Skysense \citep{guo2024skysense} extend DINO \citep{caron2021emerging} to multi-scale and multi-modal settings. SatMAE \citep{cong2022satmae}, SpectralGPT \citep{hong2024spectralgpt}, CROMA \citep{fuller2024croma}, OmniSat \citep{astruc2024omnisat}, and DOFA \citep{xiong2024neural} apply MIM for SSL 
on RS imagery. For instance, OmniSat \citep{astruc2024omnisat} introduces a modality-agnostic architecture to fuse co-registered overhead modalities (e.g., Optical, SAR, DEM) to improve representation learning. Similarly, DOFA \citep{xiong2024neural} proposes a dynamic hypernetwork to handle the heterogeneity of diverse satellite sensors. Beyond co-registered multi-sensor fusion, MMEarth \citep{nedungadi2024mmearth} constructs a global multi-modal EO corpus and shows that leveraging multiple aligned geospatial modalities can improve the learned Sentinel-2 representations. Climate FMs such as ClimaX \citep{nguyen2023climax} were pretrained on multi-source climate data using MIM. GeoVLFMs such as GeoChat \citep{kuckreja2024geochat} and EarthGPT \citep{zhang2024earthgpt} are vision-language foundation models by using aligned pairs of RS images and text, which are trained using common masked language model objectives and LoRA \citep{hu2022lora}. In parallel, DOFA-CLIP \citep{xiong2503dofa} extends language pretraining beyond RGB by aligning heterogeneous EO modalities to language with a modality-adaptive encoder Another emerging direction injects structured geo-context (e.g., OpenStreetMap) into RS pretraining; GeoLink \citep{bai2025geolink} integrates RS imagery with OSM vectors via spatially-aware cross-modal objectives to enrich representations with map-derived semantics.

However, a fundamental limitation persists in these approaches: they mainly focus on \textbf{patch-level alignment within the overhead perspective} (e.g., aligning data from different satellite sensors). 
Prominent models such as SatMAE, SpectralGPT, Skysense, CROMA, OmniSat, and DOFA typically operate under the assumption that input modalities are spatially co-registered image patches (e.g., Optical and SAR patches sharing the exact same spatial extent). 
Consequently, they fail to account for the drastic spatial scale difference between \textbf{ground-level} (street view) imagery and \textbf{overhead} imagery, as ground-level views represent a distinct, localized spatial scope that does not share a uniform extent with overhead patches. 
Furthermore, while GeoVLFMs such as GeoChat \citep{kuckreja2024geochat} align RS images with text via high-level semantics, they lack the capability to model precise \textbf{point-to-area} spatial correspondences.

\subsection{Location-Aware SSL Models}

There are several pioneering works explicitly encoding geolocation and using a contrastive learning framework between location representation and co-located image representation, such as SatCLIP \citep{klemmer2023satclip}, CSP \citep{mai2023csp},  GeoCLIP \citep{vivanco2024geoclip} and TaxaBind \citep{sastry2025taxabind}. We call them location-aware SSL models. 
By leveraging location encoding, they can explicitly model the spatial relation among different data samples. However, these frameworks either only consider a single image modality (SV imagery or RS imagery) or fail to consider the big spatial scale difference between different geospatial imagery modalities. 
Our work differs fundamentally from these works by proposing the \textbf{Geo-Aligned Implicit Representation (GAIR)} framework. By utilizing Implicit Neural Representations (INR) to model the continuous spatial domain, GAIR enables precise alignment between ground-level images and their corresponding subpixel-level locations within overhead imagery, effectively bridging the gap of dramatically different spatial scales. 



\subsection{Implicit Neural Representations}
Implicit Neural Representations (INR) have been widely applied across various domains, including image regression \citep{tancik2020fourier}, compression \citep{dupont2021coin}, 3D reconstruction \citep{mescheder2019occupancy}, image super-resolution \citep{chen2021learning,gao2023implicit}, etc. The core idea behind INR is to learn a continuous function that maps spatial coordinates to corresponding signals, enabling flexible, resolution-independent data representation. A common approach is to transform spatial coordinates into multi-scale features using Fourier feature mappings \citep{tancik2020fourier}, which are then processed by a multi-layer perceptron (MLP) to learn a continuous representation for downstream tasks. This technique has been particularly effective in image super-resolution such as LIIF \citep{chen2021learning} and CiaoSR \citep{cao2023ciaosr}, which directly learn pixel-wise feature mappings for image restoration.
In the geospatial domain, implicit neural representations have been utilized for POI type prediction \citep{mai2019space2vec}, geo-aware species fine-grained recognition \citep{mac2019presence,mai2023csp,mai2023sphere2vec,sastry2025taxabind}, species distribution modeling \citep{cole2023spatial,lange2023active,hamilton2024combining}, 
image geolocalization \citep{vivanco2024geoclip,wang2025locdiffusion}, satellite image classification and regression \citep{klemmer2023satclip,wu2024torchspatial}, and geographic question answering \citep{mai2020se,li2023location,li2022spabert}. 
In this work, 
a novel implicit neural representation module is proposed to refine the RS 
representations \( f(r_i) \) into a localized RS embedding \( \hat{r}_i(x) \) that is geographically aligned with SV 
image embedding \( g(s_i) \) and the location embedding \( e(x_i) \). 
Three embeddings -- \( \hat{r}_i(x) \), \( g(s_i) \), and \( e(x_i) \) -- are trained in a self-supervised manner through contrastive learning. 



\section{Methods}\label{method}

\subsection{Factorized Encoder for Geospatial Modalities}

\begin{figure*}[ht!]
    \centering
    \vspace{-0.5cm}
    \begin{subfigure}{1\linewidth}
        \centering
        \includegraphics[width=\linewidth]{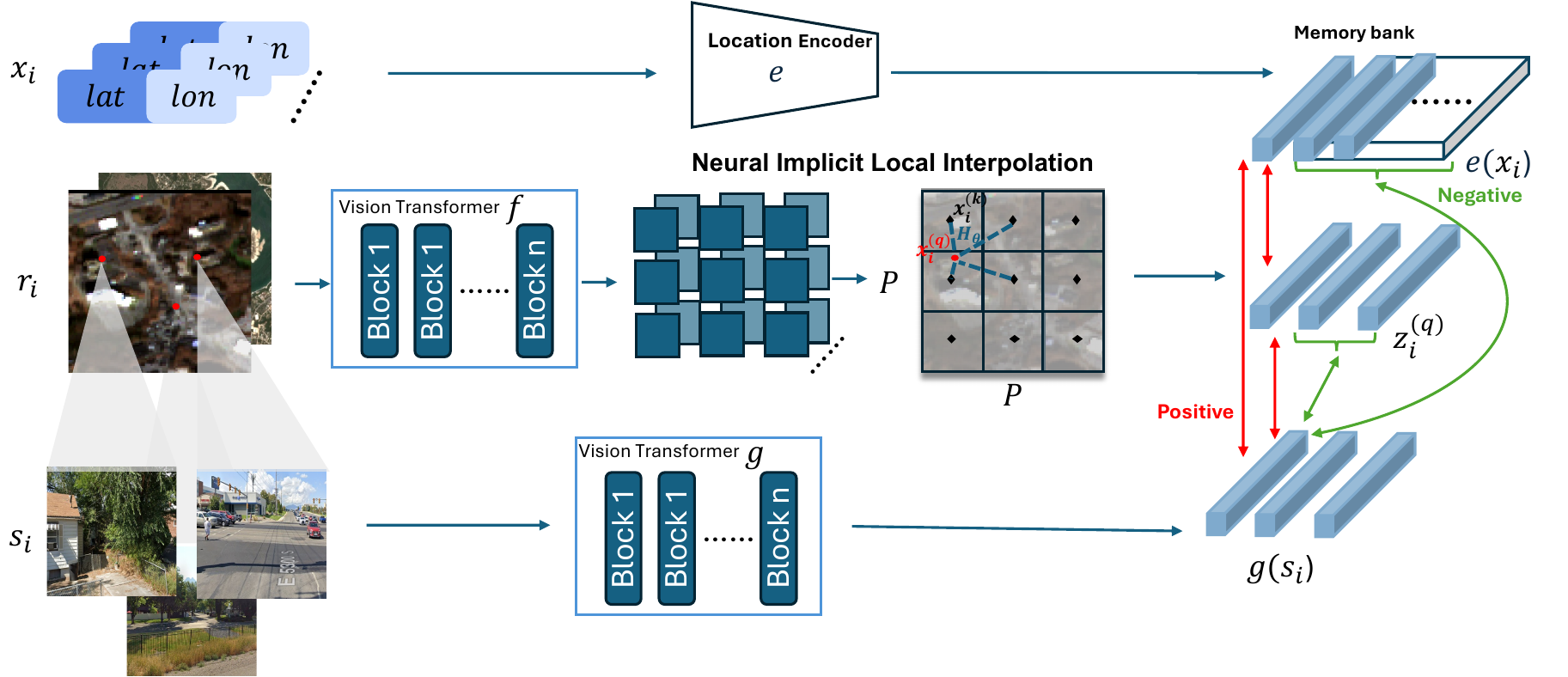}
        \caption{The \model{} Framework}
        \label{fig:model_architecture}
    \end{subfigure}
    \hfill
    \begin{subfigure}{1\linewidth}
        \centering
        \includegraphics[width=\linewidth]{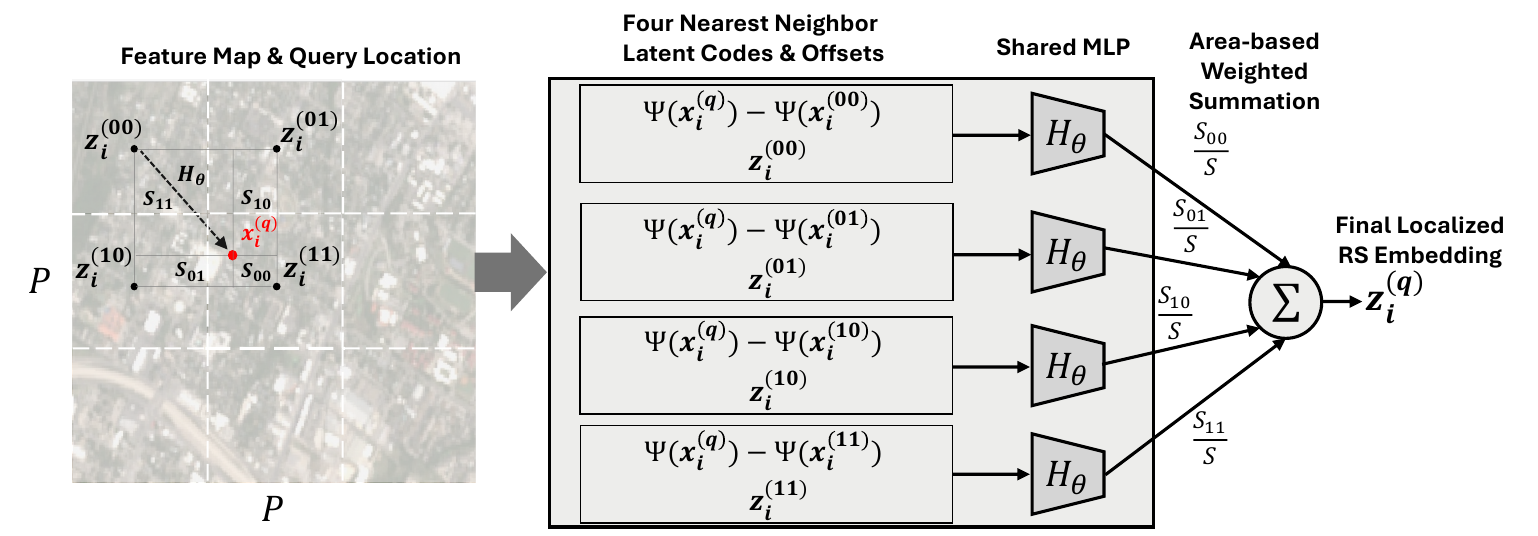}
        \caption{The NILI Module}
        \label{fig:local_interpolation}
    \end{subfigure}
    \vspace{-2mm}
    \caption{\textbf{(a) GAIR architecture:} three encoders map geolocation $x_i$, street view image $s_i$, and RS 
    image $r_i$ into embeddings. A Neural Implicit Local Interpolation (NILI) module refines $f(r_i)$ into a localized embedding $z_i^{(q)}$ at $x_i$, and contrastive learning aligns all modalities. 
    \textbf{(b) NILI illustration:} for a query location $x_i^{(q)}$, four neighboring patch embeddings $z_i^{(k)}$ are used by a shared MLP $H_\theta$ with coordinate offsets and patch embeddings from 4 nearby neighbors. The weighted sum of predictions yields the final embedding  $z_i^{(q)}$, where the weight $S_k/S$ for each reference embedding $z_i^{(k)}$ is proportional to the area of its diagonally opposite rectangle (e.g., $S_{00}$ is the area diagonally opposite to $z_i^{(00)}$). 
}
    \label{fig:fig2}
    \vspace{-0.6cm}
\end{figure*}

We define an unlabeled geo-tagged image dataset as \( X = \{(r_{i}, s_i, x_i) \mid i = 1, \dots, M \} \), where \( r_{i} \) is a remote sensing image, \( s_i \) a street view image, and \( x_i \) the location (longitude and latitude) of \( s_i \). Here, \( x_i \) is within the spatial footprint $\rsArea(r_{i})$ of \( r_{i} \) but might not be the geometric center or pixel/patch center of \( r_{i} \).
Inspired by recent contrastive pretraining models \citep{radford2021learning, mai2023csp, guo2024skysense}, we adopt a factorized encoder architecture to extract modality-specific features independently, as shown in Figure \ref{fig:fig2}. Specifically, we introduce a remote sensing image encoder \( f(\cdot) \), a street view image encoder \( g(\cdot) \), and a location encoder \( e(\cdot) \). This modular design allows each encoder to capture unique spatial and semantic characteristics with the pretrained encoder of that modality, while forming the basis of \textbf{multimodal geographically alignment} in a later stage.


\textbf{The location encoder \( e(\cdot) \)} is defined as a function \( e_{\theta}(\mathbf{x}_i) : \mathbb{S}^2 \to \mathbb{R}^d \), parameterized by \( \theta \), which maps any coordinate \( \mathbf{x}_i = (\lambda_i, \phi_i) \) on the spherical surface \( \mathbb{S}^2 \) to a \( d \)-dimensional vector representation. Here, the longitude \( \lambda_i \in [-\pi, \pi) \) and latitude \( \phi_i \in [-\pi/2, \pi/2] \). We leverage existing 2D location encoders \citep{wu2024torchspatial}, specifically the Random Fourier Features (RFF) implemented from GeoCLIP \citep{vivanco2024geoclip}, due to its strong performance in prior works. The location encoding is represented as $e(x_i)$. Accurate geolocation plays a critical role in enabling cross-modal alignment. If the coordinates are perturbed (e.g., GPS noise), the alignment between modalities becomes less reliable and the model performance degrades.We analyze this effect in Section \ref{app:noise-importance}.

\textbf{The remote sensing image encoder $ f(\cdot)$} is deployed as a Vision Transformer (ViT) \citep{dosovitskiy2020image} due to its remarkable performance and generalizability across diverse computer vision tasks in 
recent vision foundation works \citep{cong2022satmae,fuller2024croma,guo2024skysense}. Motivated by this, we employ a ViT as the RS image encoder $ f(\cdot) $. To retain a rich spatial representation for the follow-up neural implicit local interpolation operation, the output $f(r_i)$ is a patch-wise feature map, i.e., \( f(r_{i}) \in \mathbb{R}^{ P \times P \times D} \), 
where \(P^2\) is the number of patches in the ViT backbone, preserving the spatial structure of \(r_{i}\).

\textbf{The street view image encoder $ g(\cdot) $} is also a ViT model. Since we do not need the patch-level representation, we directly perform an average pooling on the patch-wise feature map to output a global image-level embedding $g(s_i)$.


\subsection{Neural Implicit Local Interpolation} \label{sec:nili}

\textbf{Since 
$s_i$'s location $x_i$ might not be at the geometric center of $r_{i}$ or any of $r_{i}$'s patch/pixel}, $g(s_i)$ and $f(r_i)$ are not directly geographically aligned 
for the contrastive learning purpose. Thus, 
we need to perform \textit{spatial interpolation} based on $f(r_{i})$ at location $x_i \in \rsArea(r_{i})$ to produce a new geo-aligned localized RS image embedding \( \rxi \). To do that, we propose a special INR module $\inrfun(\cdot)$ called \textbf{neural implicit local interpolation} which can interpolate and extract a \textbf{detailed localized RS representation} at any location $x \in \rsArea(r_{i})$.

Formally, given an overhead RS image \(r_{i}\), our RS image encoder \(f(\cdot)\) extract an image representation \( f(r_{i}) \in \mathbb{R}^{ P \times P \times D} \). 
We denote the discrete patch embeddings within $ f(r_{i}) $ as \textbf{$ \{ z_i^{(k)} \} $}, where each \textbf{$ z_i^{(k)} \in \mathbb{R}^{ D} $} corresponds to the geometric center $ x_i^{(k)} \in \rsArea(r_{i}) $ of the $ k $-th image patch. 
Here, $ k $ serves as the spatial index for the patches used in our designed INR function $\inrfun(\cdot)$.
\paragraph{Feature Unfolding for Local Context. } To enrich the patch embedding of remote sensing images with broader spatial context before interpolation, we first apply a \textbf{Feature Unfolding} strategy inspired by \citep{chen2021learning}. Specifically, for each patch embedding $z_i^{(k)}$, we concatenate it with its $3\times3$ neighboring patch embedding features. This process transforms the image representation $ f(r_{i}) \in \mathbb{R}^{ P \times P \times D} $ to $ f(r_{i}) \in \mathbb{R}^{ P \times P \times 9D} $. By incorporating information from the local neighborhood, this effectively increases the receptive field of each patch embedding, providing the INR module with the necessary information to do arbitrary location information retrieval.

\paragraph{Local Ensemble Interpolation. }
To extract a localized RS image embedding at the location $x_i$ of an SV image $s_i$, 
we develop a novel INR module $\inrfun(\cdot)$ on top of the RS image representation \( f(r_{i}) \). 
$\inrfun(\cdot)$ can learn a continuous representation of an RS image across space and be able to extract image embeddings at any query geolocation $x_{i}^{(q)} \in \rsArea(r_{i})$. 
Instead of directly retrieving a single latent code $\rxi$ at coordinate $x_{i}^{(q)}$ by using grid sample function \citep{jaderberg2015spatial}, we generate $\rxi$ by interpolating from the four nearest patch embeddings \( z_{i}^{(k)} \) (top-left, top-right, bottom-left, bottom-right) of location $x_{i}^{(q)}$ to ensure smooth transitions across space as shown in Figure \ref{fig:local_interpolation}:
\begin{align}
\rxi 
&= \inrfun(f(r_{i}), x_{i}^{(q)}) \notag \\
&= \sum_{k \in \{00,01,10,11\}} 
   \frac{S_k}{S} \cdot 
   H_\theta\!\left(
      z_i^{(k)}, 
      \Psi(x_{i}^{(q)}) - \Psi(x_i^{(k)})
   \right)
\label{eq:nili}
\end{align}

Here, \( z_i^{(k)} \in f(r_{i}) \) (\( k \in \{00,01,10,11\} \)) denote the four nearest image patch embeddings 
of query location $x_{i}^{(q)}$. 
\( x_i^{(k)} \) denotes the geographic coordinates of \( z_i^{(k)} \). $S_k$ denotes the area of the rectangle formed by the query coordinate ($x_i^{(q)}$) and the opposite corner of the grid cell relative to ($x_i^{(k)}$) (e.g., $S_{00}$ is the area of the rectangle defined by $x_i^{(q)}$ and the bottom-right corner $x_i^{(11)}$). This ensures that the weight $S_k/S$ is inversely proportional to the distance between $x_i^{(q)}$ and $x_i^{(k)}$, assigning a higher weight to the nearest patch embedding.
$\Psi(\cdot)$ is a projection function that transforms the geographic coordinates (e.g., $x_{i}^{(q)}$ and $ x_i^{(k)}$) into the image coordinate space. 
\( H_\theta \) is a multilayer perception (MLP) modulated by the latent code \( z_i^{(k)} \) which takes the projected coordinate difference between $\Psi(x_{i}^{(q)})$ and $ \Psi(x_i^{(k)})$ and predict the localized RS embedding at location $x_{i}^{(q)}$. 
The final localized RS embedding $\rxi$ is computed as the \textit{weighted sum} of the four independent predictions based on $S_k/S$ where \( S = \sum_k{S_k}\). 
This approach ensures local feature continuity by enabling overlapping representations from neighboring latent codes. At each query location, four independent predictions are ensembled, 
leading to smooth and spatially coherent feature synthesis. 
Our NILI module $\inrfun(\cdot)$ is agnostic to the implementation of $f(\cdot)$, which can be ViT- or CNN-based encoders as long as they can produce a 2D image feature map \( f(r_{i}) \in \mathbb{R}^{ P \times P \times D} \) for spatial interpolation. 

\subsection{Geo-Aligned Contrastive Objective}
Given $e(x_i)$, $\rxi$,  
and  $g(s_i)$, we can leverage their geospatial relationships 
to form SSL 
objectives to learn generalizable neural representations. 
Here, we mainly use two contrastive learning objectives: 

\paragraph{Implicit Neural Contrastive Learning (INCL). } 
The key idea of INCL is to bridge the scale discrepancy by performing contrastive learning 
between two geo-aligned image representations -- localized RS embedding $z_{i}^{(q)}$ at $x_{i}^{(q)}$ and its corresponding SV embedding $g(s_i)$:
\begin{align}
\mathcal{L}_{\text{INCL}}
&= - \frac{1}{2N}
   \sum_{i=1}^{N}
   \Bigg[
   \log
   \frac{
      \exp\!\left(\simf(\rxi, g(s_i)) / \tau\right)
   }{
      \sum\limits_{j=1}^{N}
      \exp\!\left(\simf(\rxi, g(s_j)) / \tau\right)
   } \notag \\
&\qquad\qquad
   + \log
   \frac{
      \exp\!\left(\simf(g(s_i), \rxi) / \tau\right)
   }{
      \sum\limits_{j=1}^{N}
      \exp\!\left(\simf(g(s_i), \rxj) / \tau\right)
   }
   \Bigg]
\label{eq:incl_loss}
\end{align}

%
where \( \simf(\cdot, \cdot) \) denotes a cosine similarity function, \( N \) is the batch size, and \( \tau \) is a temperature parameter. This contrastive loss enforces positive pairs between the matched embeddings while discouraging alignment with non-matching locations \citep{radford2021learning}. 
The first term ensures that the extracted localized RS embedding \( \rxi \) is aligned with the corresponding co-located SV embedding \( g(s_i) \), while the second term enforces the inverse alignment, treating \( g(s_i) \) as the anchor.

\vspace{-10pt}
\paragraph{Spatially Explicit Contrastive Learning (SECL).} 
To further reinforce the geospatial consistency across data modalities, we introduce SECL, which incorporates explicit location encoding \( e(x_i) \) into the contrastive learning framework. 
We construct a memory bank \( \mathcal{M} \) \citep{he2020momentum,vivanco2024geoclip} to store location embeddings $e(x_i)$ from past mini-batches. The SECL objective consists of two separate contrastive losses: one aligning location embeddings with RS 
embeddings, and another aligning location embeddings with SV 
embeddings. The SECL loss is defined as:
\begin{align}
\mathcal{L}_{\text{SECL}}
&= - \frac{1}{2N} \sum_{i=1}^{N}
\Bigg[
\log \frac{
\exp\!\left(\simf(e(x_i), \rxi)/\tau\right)
}{
\sum\limits_{j \in \mathcal{M}}
\exp\!\left(\simf(e(x_j), \rxi)/\tau\right)
}
\notag \\
&\qquad\qquad
+ \log \frac{
\exp\!\left(\simf(e(x_i), g(s_i))/\tau\right)
}{
\sum\limits_{j \in \mathcal{M}}
\exp\!\left(\simf(e(x_j), g(s_i))/\tau\right)
}
\Bigg]
\label{eq:secl_loss}
\end{align}

Finally, \model{}'s pre-training objective is the sum of two objectives, where $\lambda$ is a hyperparameter to control the contribution of SECL:
\begin{equation}
    \mathcal{L} = \mathcal{L}_{\text{INCL}} + \lambda\mathcal{L}_{\text{SECL}}
\end{equation}

\subsection{Transfer Learning}
To assess the performance of the pre-trained \model{}, we employ it both as a feature extractor for linear probing and as a model parameter initialization for model fine-tuning. \model{} comprises three encoders: two image encoders dedicated to remote sensing and street view images, and a location encoder responsible for location data representation.
\paragraph{Fine-Tuning the StreetView Image Encoder $g(\cdot)$.} For the SV image encoder $g(\cdot)$, we remove the projection layer and introduce a new head \( h_{g}(\cdot) \) to process the extracted SV image feature vectors. 
We adopt different experimental setups --
\( h_{g}(\cdot) \) is a single linear layer for linear probing and two linear layers for non-linear probing. 
\paragraph{Fine-Tuning the Remote Sensing Image Encoder $f(\cdot)$.}
Similarly, we remove the projection layer and introduce a new head \( h_{f}(\cdot) \). 
For RS image semantic segmentation tasks, we use UPerNet \citep{xiao2018unified} as \( h_{f}(\cdot) \), while for RS image change detection, we adopt a Siamese UPerNet as \( h_{f}(\cdot) \) to capture temporal changes.  For multi-temporal datasets, we experiment with two temporal aggregation strategies: (1) a naive linear mapping, and (2) a lightweight spatial-temporal encoder (L-TAE) \citep{garnot2020lightweight} to better integrate temporal information.
\paragraph{Fine-Tuning the Location Encoder $e(\cdot)$.} We also introduce a new head \( h_e(\cdot) \) for the location encoder \( e(\cdot) \). Following standard protocols for the geo-aware image classification \citep{mac2019presence,wu2024torchspatial}, given a location-image pair \( (x, I) \), where \( I \) denotes an image, the model predicts the category \( y \) by factorizing the probability as \( P(y|I, x) \propto P(y|I) P(y|x) \). Here, \( P(y|I) \) is estimated using an image encoder; in this paper, we utilize Inception V3 network \citep{szegedy2016rethinking} following the standard implementation \citep{mai2019space2vec,wu2024torchspatial}, while \( e(x) \) contributes to \( P(y|x) \). The location encoder is fine-tuned using cross-entropy loss for the geo-aware image classification tasks. For geo-aware image regression tasks, we concatenate the image embedding and location embedding and feed the result into a linear layer for regression. We also use MSE as the loss function. 


\section{Experiments}
\label{exp}
\subsection{Implementation Details}
\subsubsection{Pretraining Dataset Construction.} 
\begin{figure*}[ht!]
    \centering
    \includegraphics[width=1\linewidth]{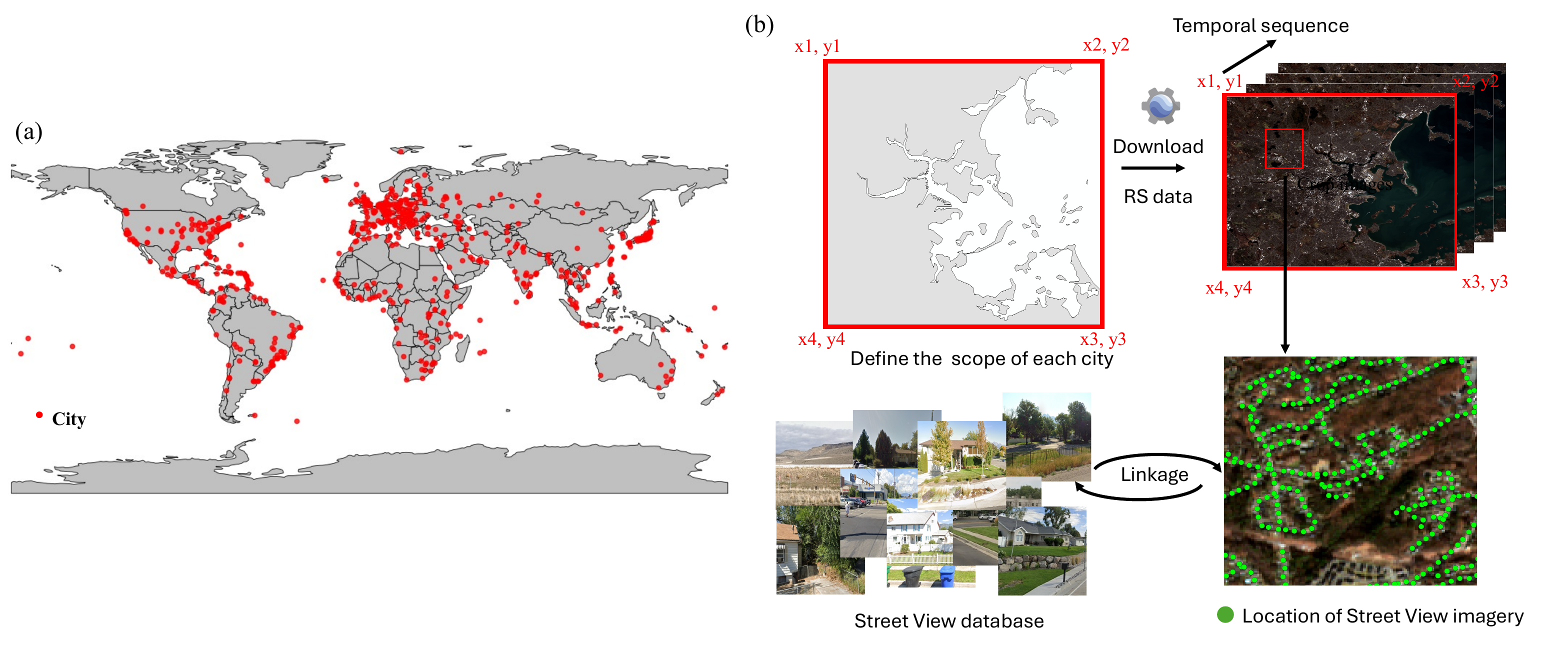}  
    \caption{An illustration of the pipeline used to construct our \textbf{Streetscapes1M} GeoFM pretraining dataset. (a) The geographic distribution of the 688 cities included in the dataset. (b) The three-step construction pipeline: study area selection, RS data collection, and \textbf{multimodal data pairing}.} 
    \label{fig:data_process}
\end{figure*} 

To facilitate the pre-training of the \model{} model, we construct a global-scale, geospatially aligned multimodal dataset named \textbf{Streetscapes1M}, derived from the Global Streetscapes dataset \citep{hou2024global}. The construction pipeline, illustrated in Figure \ref{fig:data_process}, consists of three primary stages:

\begin{enumerate}
    \item \textbf{Study Area Selection:} We define a diverse study area encompassing 688 cities globally (Figure \ref{fig:data_process}a), utilizing the city boundary definitions (bounding boxes) provided by the Global Streetscapes dataset. From this collection, we sample 1 million street view images to serve as the foundation of our dataset. To ensure a balanced geographic distribution, we employ a random uniform sampling strategy, selecting approximately 1,453 street view images from each of the 688 cities.
    
    \item \textbf{Remote Sensing (RS) Image Collection:} For each sampled location, we define the spatiotemporal scope to retrieve Sentinel-2 multispectral imagery via Google Earth Engine (GEE). We acquire all available imagery from 2017 to 2024, filtering for Level-2A atmospherically corrected products with less than 20\% cloud cover. Following established protocols for RS foundation models \citep{cong2022satmae, fuller2024croma}, we utilize 10m spatial resolution bands along with 20m bands upsampled to 10m, and exclude the Coastal Aerosol (B1), Water Vapor (B9), and cirrus band (B10). Consequently, all retained bands are unified to 10m resolution via resampling to minimize noise. To ensure diversity in surface conditions while controlling for cloud contamination, we retrieve a time series of qualifying monthly images and apply random temporal selection during training which acts as satellite image temporal augmentation by following existing literature \citep{ayush2021geography,manas2021seasonal}.
    
    \item \textbf{Multimodal Data Pairing:} Based on the exact geolocation of each street view image, we retrieve and pair it with the corresponding overhead RS imagery that spatially covers these coordinates. This process generates the matched multimodal triples $(r_i, s_i, x_i)$ required for \model{} pre-training, ensuring that each ground-level view is located within its paired satellite image footprint.
\end{enumerate}

\subsubsection{Implementation Details.} 
The \textbf{Streetscapes1M} dataset comprises 1 million triples $(r_i, s_i, x_i)$, where $r_i$ denotes the remote sensing image, $s_i$ the street view image, and $x_i$ the geolocation. All models, including baselines and \model{}, are implemented using a ViT-B backbone. Specifically, the remote sensing encoder employs a smaller patch size of $8 \times 8$ to better capture fine-grained textures. In line with standard practices in multimodal foundation model pretraining \citep{klemmer2023satclip,liu2024remoteclip,vivanco2024geoclip,xiong2024neural}, the remote sensing encoder is initialized with the pretrained optical checkpoint from CROMA \citep{fuller2024croma}, while all other network components are trained from scratch.

During training, we fix the input resolution of street view images to $224 \times 224$ and remote sensing images to $96 \times 96$; handling dynamic input sizes is left for future work. We apply two types of data augmentations: (1) standard augmentations for both modalities, including random horizontal flipping and color jittering, and (2) temporal augmentation for Sentinel-2 imagery, where a timestamp is randomly selected from the time series following \cite{ayush2021geography}.

We train our model on a Linux server equipped with 4 NVIDIA RTX A6000 GPUs (48GB) using a batch size of 256. Optimization is performed using AdamW with $\beta_1=0.9$, $\beta_2=0.999$, and a weight decay of 0.01. The base learning rate is set to $1.5 \times 10^{-6}$ with a linear warm-up for the first 5\% of epochs, and the loss balancing parameter $\lambda$ is set to 0.5. 

\begin{figure*}[t]
    \centering
    \includegraphics[width=0.85\linewidth]{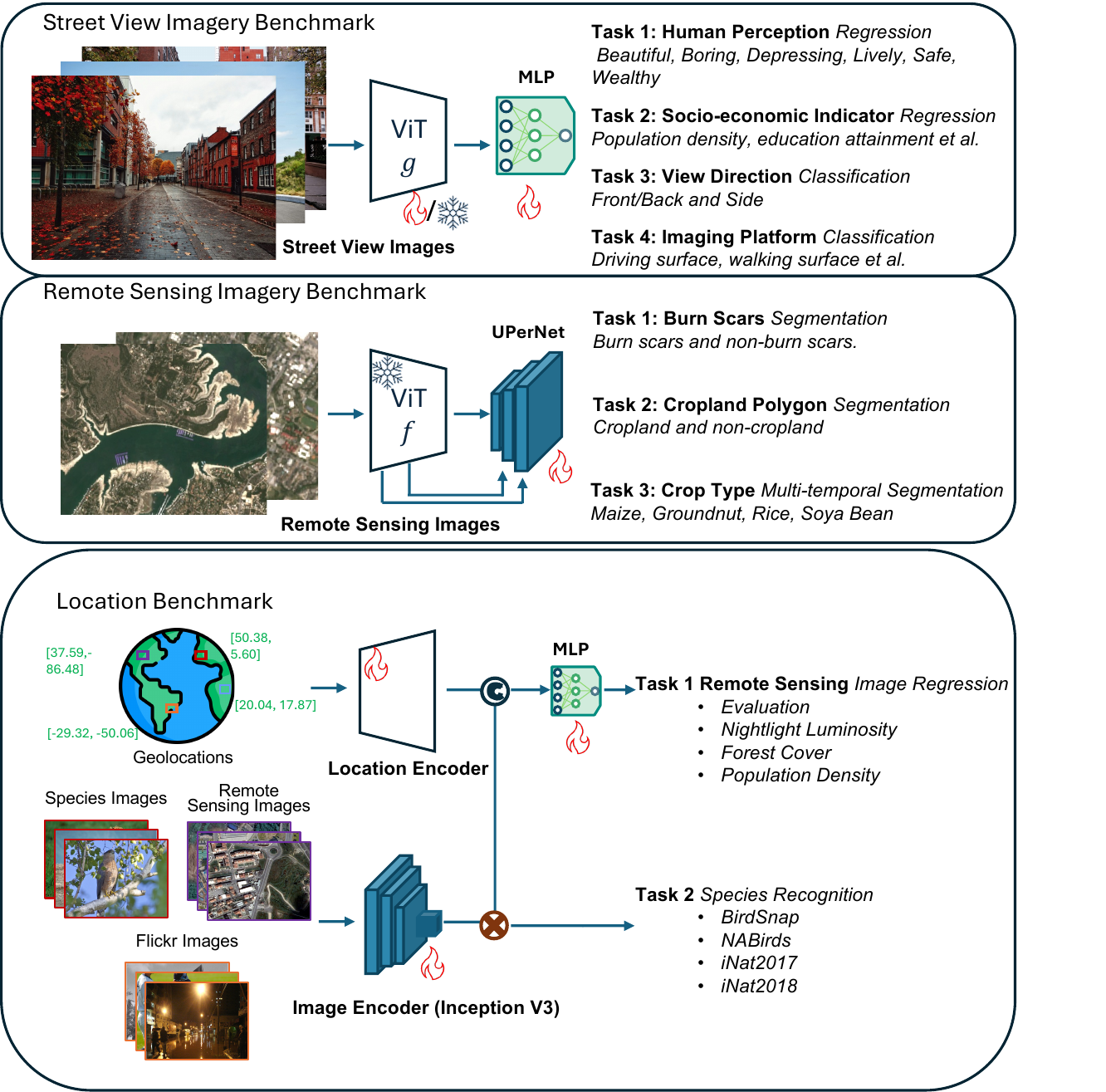} 
    \caption{
    The  evaluation pipelines of three benchmarks
    (9 tasks). 
    After pretraining \model, we fine-tune the street view image encoder \( g \) to tackle the street view imagery benchmark, the RS image encoder \( f \) for remote sensing imagery benchmark, and the location encoder \( e \) for location benchmark.}
    \vspace{-0.6cm}
    \label{fig:evaluation}
\end{figure*} 

\subsubsection{Model Fine-Tuning and Evaluation.} 
We evaluate \model{} across three distinct benchmarks: street view imagery, remote sensing imagery, and location-based tasks. A comprehensive overview of the fine-tuning pipeline, baselines, and task configurations is provided in Figure \ref{fig:evaluation}. Unless otherwise specified, all baselines follow a standardized setup with a batch size of 256, 50 training epochs, and a learning rate of $1 \times 10^{-3}$. Images are resized to match the specific input requirements of each baseline model.

\textbf{Street View Imagery.} Due to the lack of official implementations, we reimplement all street view baselines. We adopt three strategies to adapt the model: 
1) \textit{Linear probing}: The encoder remains frozen, and only a linear head (single-layer MLP) is trained; 
2) \textit{Non-linear probing}: The encoder is fixed, and a two-layer MLP head with a sigmoid activation function is optimized; 
3) \textit{Full fine-tuning}: All model parameters are updated.

\textbf{Remote Sensing Imagery.} Following the protocols of PANGAEA-Bench \citep{marsocci2024pangaeaglobalinclusivebenchmark}, we evaluate representation quality under two settings: freezing the pretrained backbone to fine-tune only the UPerNet head, and fully fine-tuning the entire model.

\textbf{Location-based Tasks.} We utilize baselines from TorchSpatial \citep{wu2024torchspatial}. Due to the lightweight nature of the location encoder, we perform full fine-tuning on all parameters initialized with our pretrained weights.
\subsection{Benchmarks and Baselines}
\label{sec:benchmark_baseline}

To comprehensively evaluate \model{}, we conduct experiments on 9 geospatial tasks across 22 datasets. These tasks are categorized into three domains: street view imagery, remote sensing imagery, and location-based tasks. Figure \ref{fig:evaluation} provides a schematic overview.

\subsubsection{Benchmarks}
\paragraph{Street View Imagery Tasks.}
We utilize ground-level imagery to predict socio-economic indicators, human perception metrics, and image metadata. We partition all datasets into a 60\%/20\%/20\% (train/val/test) split using an intra-city random strategy to ensure proportional city representation. Specifically, the socio-economic regression uses imagery from Los Angeles and Boston \citep{fan2023urban}, while the remaining tasks use the Global Streetscapes dataset \citep{hou2024global}.
\begin{itemize}
    \item \textbf{Socio-Economic Indicator Regression:} We predict ten socio-economic indicators using 410,286 street view images from Los Angeles and Boston \citep{fan2023urban}. Indicators include population density, educational attainment, health conditions, racial demographics, median income, public transport proximity, active mobility rates, age demographics, crime rates, and visible sky area.
    \item \textbf{Human Perception Regression:} Based on the Global Streetscapes dataset \citep{hou2024global}, this task predicts human perception ratings (ranging 0-1) for six attributes: 'Beautiful', 'Boring', 'Depressing', 'Lively', 'Safe', and 'Wealthy'.
    \item \textbf{View Direction Classification:} A classification task using Global Streetscapes labels to determine if an image faces "Front/Back" or "Side".
    \item \textbf{Imaging Platform Classification:} Classifies the capture platform into 6 categories: driving surface, walking surface, cycling surface, tunnel, open fields, and railway.
\end{itemize}

\paragraph{Remote Sensing Tasks.}
These benchmarks evaluate performance on various overhead imagery tasks. To ensure fair comparison with existing baselines, we strictly adhere to the official training, validation, and testing splits provided by the respective benchmarks.
\begin{itemize}
    \item \textbf{Burn Scars Estimation:} Uses the HLS Burn Scars dataset \citep{jakubik2023foundation} (804 scenes, $512 \times 512$ pixels) covering the US (2018-2021). Input includes six Sentinel-2 spectral bands (Blue, Green, Red, NIR, SW1, SW2) to segment burn scar masks.
    \item \textbf{Crop Type Mapping (South Sudan):} Based on \citet{m2019semantic}, this task segments 4 crop types using Sentinel-2 imagery ($64 \times 64$ pixels, 10 bands) across 837 agricultural fields.
    \item \textbf{Cropland Boundary Delineation:} Uses the AI4Small -Farms dataset \citep{persello2023ai4smallfarms} containing 439,001 manually annotated field polygons. Models utilize four spectral bands (B2, B3, B4, B8) for delineation.
\end{itemize}

\paragraph{Location-Based Tasks.}
These tasks incorporate geolocation metadata alongside imagery to evaluate geospatial representation learning. Similar to the remote sensing tasks, we adopt the standard data partitions defined in the original datasets.
\begin{itemize}
    \item \textbf{Geo-Aware Image Regression (MOSAIKS):} We use the MOSAIKS dataset \citep{rolf2021generalizable} to predict geospatial variables from satellite imagery. Specific tasks include: 
    (1) \textit{Population Density}: 425,637 samples, log-transformed; 
    (2) \textit{Forest Cover}: 498,106 observations of vegetation $>5$m height; 
    (3) \textit{Nightlight Luminosity}: 492,226 observations based on VIIRS data; 
    (4) \textit{Elevation}: 498,115 records derived from SRTM data.
    
    \item \textbf{Geo-Aware Species Recognition:} Classifies fine-grained species categories using images and location metadata. Datasets include: 
    (1) \textit{BirdSnap} \citep{berg2014birdsnap}: 19,567 images, 500 species; 
    (2) \textit{NABirds} \citep{van2015building}: 23,699 images, 555 species; 
    (3) \textit{iNat2017} \citep{van2018inaturalist}: 675,170 images, 5,089 species; 
    (4) \textit{iNat2018}: 461,939 images, 8,142 species.
\end{itemize}

\subsubsection{Baselines}

We select a diverse set of baselines for each benchmark category to ensure a fair and comprehensive comparison.

\paragraph{Street View Imagery Baselines.}
We compare \model{} against three categories of models:
\begin{itemize}
    \item \textbf{Geo-Foundation Models (GeoFMs):} We include \textbf{SatMAE} \citep{cong2022satmae} (masked autoencoder for RS), \textbf{CROMA} \citep{fuller2024croma} (multimodal contrastive learning), \textbf{PIS} \citep{an2024pretrain} (intra-instance similarity), and the ground image encoder from \textbf{TaxaBind} \citep{sastry2025taxabind}.
    \item \textbf{General Vision Models:} We benchmark against standard ViTs with \textbf{Random Initialization} and \textbf{ImageNet Initialization} (supervised).
    \item \textbf{Self-Supervised Learning (SSL) Models:} We include ViTs pretrained with \textbf{MoCo V3} and \textbf{MAE} on ImageNet. To analyze the benefit of domain adaptation, we also include \textbf{MoCo V3-Streetscapes} and \textbf{MAE-Streetscapes}, which are trained directly on our Streetscapes1M dataset.
\end{itemize}

\paragraph{Remote Sensing Imagery Baselines.}
We adopt \textbf{PANGAEA-Bench} \citep{marsocci2024pangaeaglobalinclusivebenchmark}, which includes 12 established GeoFM baselines. Additionally, we include the recent representative multi-model satellite encoder from TaxaBind \citep{sastry2025taxabind}, GeoLink \citep{bai2025geolink}, DOFA-CLIP \citep{xiong2503dofa}, MMEarth and MMEarth64 \citep{nedungadi2024mmearth}.

\paragraph{Location-Based Baselines.}
We employ \textbf{LocBench} from TorchSpatial \citep{wu2024torchspatial} to evaluate different Random Fourier Feature (RFF) initialization strategies:
1) \textbf{Random Init.} (supervised training on downstream tasks without pretraining);
2) \textbf{GeoCLIP Init.} \citep{vivanco2024geoclip};
3) \textbf{TaxaBind Init.} \citep{sastry2025taxabind} (using the TaxaBind location encoder);
4) \textbf{\model{} Init.} (ours).
We also include a \textbf{No Prior} baseline, which uses only image features, omitting location embeddings entirely.
\paragraph{Ablation Study.}
We conduct ablation studies to analyze the impact of different components in \model{}, including six variants:

\begin{itemize}[leftmargin=12pt]
\setlength\itemsep{-0.2em}
\item \textbf{\model{}-MAE} – In this setting, we replace \model{}'s contrastive learning objective with a masked autoencoder (MAE) objective \citep{he2022masked}. Specifically, we do masking on the RS image and the street view images are treated as a special masked patch of the RS image, allowing the model to reconstruct missing RS image patches.
\item \textbf{\model{} w/o Loc} – To assess the significance of geolocation encoding, we remove the location embedding $ e(x_i) $ from \model{}. 
\item \textbf{\model{} w/o NILI}: This variant removes the implicit neural representation (INR) module for spatially aligning remote sensing features with street view images. Instead, we apply global average pooling over the feature map produced by the remote sensing encoder $f(r_i)$ to obtain a single latent vector. This representation is then used directly for contrastive learning without the geospatial alignment step. 
\item \textbf{\model{} w/o NILI and Loc}: In this setting, both the INR module and the location encoder branch are removed. The model performs contrastive learning only between remote sensing and street view embeddings. This baseline is conceptually similar to \citep{huynh2025contrastive}. The difference is that \citet{huynh2025contrastive} used the embeddings of species images instead of streetview images to contrast with the remote sensing image embeddings.
\item \textbf{\model{}-Bilinear}: In this setting, we replace the learnable NILI module with standard bilinear interpolation. Instead of using the INR-based generation, we directly sample the localized embedding from the remote sensing feature map $f(r_i)$ at coordinate $x_i$ via bilinear interpolation.
\item \textbf{\model{}-Bicubic}: Similarly, we replace the learnable NILI module with standard bicubic interpolation. Both \model{}-Bilinear and \model{}-Bicubic serve to validate the effectiveness of the proposed neural implicit interpolation against deterministic interpolation methods.

\end{itemize}


\subsection{Street View Imagery Results}
\begin{table*}[t!]
    \centering
    \renewcommand{\arraystretch}{1.0}
    \setlength{\tabcolsep}{6pt} 
    \caption{Comparisons of model performance across street view imagery benchmark. RMSE $\downarrow$ is reported for \textit{socio-economic indicator regression and human perception regression}, while F1 score $\uparrow$ is reported for \textit{view direction classification and imaging platform classification}. The best and second-best results are denoted as \textbf{Bold} and \underline{Underline}.}
    \label{tab:street_view_benchmark}
    \setlength{\tabcolsep}{3pt}
    \resizebox{\textwidth}{!}{%
    \begin{tabular}{lccc|ccc||ccc|ccc}
        \toprule
        \multirow{2}{*}{Model} & \multicolumn{3}{c|}{Socio-Eco. Indic. (RMSE $\downarrow$)} & \multicolumn{3}{c||}{Human Perception (RMSE $\downarrow$)} & \multicolumn{3}{c|}{View Direction (F1 $\uparrow$)} & \multicolumn{3}{c}{Imaging Platform (F1 $\uparrow$)} \\
        & Linear & Non-Linear & All & Linear & Non-Linear & All & Linear & Non-Linear & All & Linear & Non-Linear & All \\
        \midrule
        SatMAE\citep{cong2022satmae}  & 0.9671  & 0.9421  & 0.7690  & 2.1681  & 2.1196  & 1.8812  & 0.0000  & 0.0202  & 0.3885  & 0.1916  & 0.1596  & 0.2114 \\
        CROMA\citep{fuller2024croma}  & 0.9550  & 0.9295  & 0.7927  & 2.0768  & 2.0219  & 1.9387  & 0.0064  & 0.0640  & 0.2133  & 0.1733  & 0.1832  & 0.2369 \\
        PIS\citep{an2024pretrain}     & 0.9290  & 0.8911  & 0.6797  & 1.9858  & 1.8865  & 1.6166  & 0.1937  & 0.2442  & 0.3537  & 0.2315  & 0.2796  & 0.3328 \\
        TaxaBind\citep{sastry2025taxabind} & 0.8819 & \textbf{0.8296} & \underline{0.6654} & 1.7960 & 1.7108 & 1.6067 & 0.2617 & 0.3971 & 0.5088 & 0.2323 & 0.2541 & 0.3052 \\
        \midrule
        Random Init.        & 0.9700  & 0.9524  & 0.9035  & 2.2000  & 2.1534  & 2.1839  & 0.0000  & 0.0000  & 0.3954  & 0.1834  & 0.1895  & 0.1970 \\
        ImageNet Init.\citep{wu2020visual} & 0.8929  & 0.8489  & 0.7012  & \underline{1.7473}  & \textbf{1.6439}  & \underline{1.5271}  & \underline{0.5328}  & 0.4186  & 0.5850  & 0.3229  & 0.3275  & 0.3540 \\
        MoCo V3
        \citep{chen2021mocov3}  & \textbf{0.8760}  & 0.8441  & 0.6779  & 1.7590  & 1.6599  & 1.5821  & 0.2106  & 0.6262  & 0.4976  & 0.2584  & 0.3171  & 0.3161 \\
        MAE-ImageNet\citep{he2022masked} & 0.8816  & 0.8373  & 0.7150  & 1.8730  & 1.7592  & 1.5788  & 0.4298  & \underline{0.6426}  & 0.4961  & 0.2619  & 0.3257  & 0.2195 \\
        \midrule
        MoCo V3-Streetscapes  & 0.8863  & 0.8429  & 0.6707  & 1.7601  & 1.7084  & 1.5800  & 0.3508  & 0.5988  & 0.5032  & 0.3439  & 0.3308  & 0.3639 \\
        MAE-Streetscapes      & 0.8940  & 0.8416  & 0.6872  & 1.8514  & 1.7701  & 1.5774  & 0.3451  & 0.6102  & 0.4921  & 0.3335  & 0.3482  & 0.3050 \\
        \midrule
        \model{}-MAE          & 0.9485  & 0.9011  & 0.8048  & 2.1254  & 2.0495  & 1.9611  & 0.0272  & 0.0895  & 0.3029  & 0.2124  & 0.2582  & 0.3249 \\
        \model{} w/o Loc      & 0.8823  & 0.8352  & 0.6725  & \underline{1.7473}  & 1.6782  & 1.5960  & 0.5290  & 0.6052  & \underline{0.6078}  & \underline{0.3676}  & \underline{0.3709}  & \underline{0.3892} \\
        \model{}              & \underline{0.8803}  & \underline{0.8349}  & \textbf{0.6612}  & \textbf{1.7141}  & \underline{1.6489}  & \textbf{1.5072}  & \textbf{0.5457}  & \textbf{0.6495}  & \textbf{0.6102}  & \textbf{0.3793}  & \textbf{0.3782}  & \textbf{0.4071} \\
        \bottomrule
    \end{tabular}}

\end{table*}

\paragraph{Socio-economic Indicator Regression.} This task aims at predicting socio-economic indicators from street view images, a widely studied problem in urban analytics \citep{fan2023urban}. We evaluate models on 10 socio-economic indicators: population density, educational attainment, health condition rates, racial demographics, median household income, proximity to public transportation, percentage of people who walk or bike, proportion of the population over 65 years old, crime rates, and visible sky area, and report RMSE across different fine-tuning settings in Table \ref{tab:street_view_benchmark}. \model{} achieves the lowest RMSE in the Fine-Tune All settings and the 2nd best in Non-Linear Probing settings. 
\paragraph{Human Perception Regression.} This task assesses human perceptions of urban environments, which are widely used in urban planning, psychology, and social studies \citep{wei2022mapping}. We perform regression on six perceptual attributes: 'Beautiful', 'Boring', 'Depressing', 'Lively', 'Safe', and 'Wealthy'. As shown in Table \ref{tab:street_view_benchmark}, we can see that \model{} achieves the best performance on linear probing and fine-tuning settings, and remains the 2nd best model 
on the non-linear probing setting. 
Notably, the explicit integration of geolocation into contrastive learning significantly enhances model performance, as seen in the substantial improvement from \model{} w/o Loc to \model{}.
\paragraph{View Direction Classification.} View direction reflects the model's ability to capture geospatial context. This task involves classifying images into two categories: "front/back" and "side". As shown in Table \ref{tab:street_view_benchmark}, existing GeoFMs struggle with view direction estimation, whereas \model{} consistently outperforms all baselines, demonstrating its superior spatial awareness.
\paragraph{Imaging Platform Classification.} Different imaging platforms offer distinct perspectives on the built and natural environment. For this task, we include 6 platform types, including driving surface, walking surface, cycling surface, tunnel, open fields, and railway. As shown in Table \ref{tab:street_view_benchmark}, \model{} outperforms all baselines on all settings.

\begin{table*}[t]
\caption{Performance comparison (mIoU $\uparrow$) of \model{} and other Geo-Foundation Models (GeoFMs) across four remote sensing semantic segmentation (SS) tasks. We report results under two protocols: tuning head only and full fine-tuning. ``Linear'' and ``L-TAE'' under Crop Type represent two multi-temporal aggregation strategies.}
\label{tab:remote_sensing_benchmark}
\centering
\setlength{\tabcolsep}{4pt}
\resizebox{0.9\textwidth}{!}{%
\begin{tabular}{l l cccc cccc}
\toprule
\multirow{3}{*}{\textbf{Model}} & \multirow{3}{*}{\textbf{RS Encoder Config}} & \multicolumn{4}{c}{\textbf{Fine-tuning Head}} & \multicolumn{4}{c}{\textbf{Full Fine-tuning}} \\
\cmidrule(lr){3-6} \cmidrule(l){7-10}
 &  & \multirow{2}{*}{Burn Scars} & \multirow{2}{*}{Crop. Poly.} & \multicolumn{2}{c}{Crop Type} & \multirow{2}{*}{Burn Scars} & \multirow{2}{*}{Crop. Poly.} & \multicolumn{2}{c}{Crop Type} \\
\cmidrule(lr){5-6} \cmidrule(l){9-10}
 &  &  &  & Linear & L-TAE &  &  & Linear & L-TAE \\
\midrule
CROMA \citep{fuller2024croma}       & ViT-L/8            & 81.95 & 25.65 & 47.02 & 49.38 & 84.40 & 37.84 & 41.15 & 32.11 \\
DOFA \citep{xiong2024neural}        & ViT-B/16         & 78.96 & 27.07 & 49.81 & 51.33 & 87.36 & 37.84 & 51.00 & 46.67 \\
GFM-Swin \citep{mendieta2023gfm}    & Swin-B           & 76.17 & 27.19 & 39.72 & 46.98 & 79.39 & 26.03 & 28.73 & 27.48 \\
Prithvi \citep{jakubik2023foundation} & ViT-B/16       & 82.67 & 26.86 & 39.92 & 43.07 & 81.41 & 38.83 & 34.92 & 43.16 \\
RemoteCLIP \citep{liu2024remoteclip}& ViT-B/32         & 75.55 & 25.12 & 46.50 & 52.05 & 72.27 & 37.71 & 22.78 & 28.93 \\
SatlasNet \citep{bastani2023satlaspretrain} & Swin-B   & 79.69 & 25.13 & 46.97 & 46.97 & 85.21 & 26.34 & 47.31 & 53.43 \\
Scale-MAE \citep{reed2023scale}     & ViT-L/16         & 76.71 & 21.47 & 21.39 & 25.42 & 77.52 & 38.00 & 19.66 & 34.71 \\
SpectralGPT \citep{hong2024spectralgpt}& ViT-B/8       & 80.47 & 26.75 & 53.50 & 46.95 & 82.17 & 33.61 & 51.28 & 57.65 \\
TaxaBind \citep{sastry2025taxabind} & ViT-B/16         & 75.84 & 38.46 & 44.80 & 43.52 & 75.88 & 36.60 & 46.26 & 29.29 \\
S12-Data2Vec \citep{stewart2023ssl4eo} & ViT-S/16      & 81.14 & 24.23 & 54.01 & 54.03 & 82.89 & 35.90 & 52.51 & 50.51 \\
S12-DINO \citep{stewart2023ssl4eo}  & ViT-S/16         & 81.44 & 25.62 & 46.56 & 48.66 & 84.36 & 35.40 & 53.07 & 55.89 \\
S12-MAE \citep{stewart2023ssl4eo}   & ViT-S/16         & 80.86 & 24.69 & 46.28 & 45.80 & 83.80 & 32.67 & 51.76 & 48.58 \\
S12-MoCo \citep{stewart2023ssl4eo}  & ViT-S/16         & 80.76 & 25.38 & 44.22 & 48.58 & 84.15 & 34.40 & 50.72 & 51.56 \\
MMEarth \citep{nedungadi2024mmearth} & ConvNeXtV2-Atto & 81.00 & 37.85 & 34.00 & 33.53 & 82.51 & 37.85 & 32.04 & 53.79 \\
MMEarth64 \citep{nedungadi2024mmearth}& ConvNeXtV2-Atto & 79.58 & 35.10 & 52.69 & 51.06 & 82.48 & 32.82 & 54.59 & 51.65 \\
DOFA-CLIP \citep{xiong2503dofa} & ViT-SO400M/14 & 80.40 & 32.21 & 52.22 & 49.10 & \textbf{92.13} & 32.74 & 44.77 & 42.39 \\
GeoLink \citep{bai2025geolink}  & ViT-L/16 & 76.33 & 33.18 & 36.82 & 28.09 & 80.41 & 37.70 & 40.42 & 21.59 \\
\midrule
\model{}-MAE                        & ViT-B/8          & 74.15 & 22.77 & 34.18 & 40.44 & 76.18 & 37.12 & 45.51 & 40.43 \\
\model{} w/o Loc                    & ViT-B/8          & 82.94 & 43.28 & 55.41 & \underline{54.32} & 86.14 & 41.88 & 54.92 & \textbf{58.12} \\
\model{}                            & ViT-B/8          & \underline{83.26} & \underline{43.35} & \underline{55.53} & 54.01 & \underline{87.00} & \underline{42.51} & \underline{55.66} & 57.90 \\
\model{} \textit{Debias}     & ViT-B/8   & \textbf{83.78} & \textbf{43.47} & \textbf{55.70} & \textbf{54.51} & 86.88 & \textbf{42.88} & \textbf{55.87} & \underline{57.92} \\
\bottomrule
\end{tabular}%
}
\end{table*}

\subsection{Remote Sensing Results}  
Table \ref{tab:remote_sensing_benchmark} presents the evaluation results of \model{} across three remote sensing tasks.

\paragraph{Single Temporal Semantic Segmentation. } We evaluate \model{} on two single-temporal segmentation tasks: burn scar segmentation using the HLS Burns dataset \citep{jakubik2023foundation} and cropland polygon delineation using the AI4SmallFarms dataset \citep{persello2023ai4smallfarms}, both based on Sentinel-2 imagery. In fine-tuning head setting, \model{} achieves a mean Intersection over Union (mIoU) of 83.26\% for burn scars and 43.35\% for cropland segmentation, outperforming all baselines. Notably, \model{} improves performance by 0.5\% for burn scars and 16.16\% for cropland segmentation, demonstrating its capability in extracting meaningful geospatial features for land cover classification. Furthermore, in the full fine-tuning setting, \model{} continues to achieve highly competitive results, with 87.00\% for burn scars and 42.51\% for cropland segmentation. It is noteworthy that \model{} utilizes a standard ViT-B architecture, yet it consistently matches or significantly outperforms larger-capacity foundation models, including CROMA (ViT-L), Scale-MAE (ViT-L), and GeoLink (ViT-L). Although \model{} underperforms the massive DOFA-CLIP (ViT-SO400M) on Burn Scars, this asymmetric comparison underscores our model's exceptional parameter efficiency.

\paragraph{Multi Temporal Semantic Segmentation. } For this task, we utilize the crop type mapping dataset \citep{m2019semantic}, which consists of Sentinel-2 imagery from 2017. To utilize temporal information, we employ two widely used feature aggregation strategies: linear aggregation and Lightweight Temporal Attention Encoder (L-TAE) aggregation \citep{garnot2020lightweight}. In fine-tuning head setting, model{} consistently outperforms other baselines, particularly in the simpler linear aggregation setting, achieving a 1\% mIoU improvement. This superiority extends to the full fine-tuning setting, where \model{} achieves 55.66\% and 57.90\% mIoU for linear and L-TAE aggregation, respectively. These results underscore that our cross-view alignment strategy effectively scales to multi-temporal visual sequences, allowing our ViT-B model to consistently surpass other ViT-B scale competitors (e.g., DOFA, Prithvi, SpectralGPT) and larger settings in complex temporal modeling tasks.

\begin{table*}[ht!]
\centering
\caption{The location benchmark results (LocBench)  
for image regression 
and species recognition. The ``PopDen'', ``ForCov'', ``NightLum'', and ``Elev.'' columns indicate regression prediction results on population density, forest coverage, nightlight luminosity, and elevation.
} 
\resizebox{0.99\textwidth}{!}{%
\begin{tabular}{llcccccccc}
\toprule
Init. & Model & \multicolumn{4}{c}{Image Regression (R$^2$ $\uparrow$)} & \multicolumn{4}{c}{Species Recognition (Top-1 accuracy $\uparrow$)} \\
\cmidrule(lr){3-6} \cmidrule(lr){7-10}
 &  & PopDen & ForCov & NightLum & Elev. & BirdSnap & NABirds & iNat17 & iNat18 \\
\midrule
\multirow{2}{*}{Rand \citep{wu2024torchspatial}} 
    & No Prior & 0.38 & 0.52 & 0.33 & 0.27 & 70.07 & 76.08 & 63.27 & 60.20 \\
    & RFF      & 0.57 & \underline{0.84} & 0.35 & 0.76 & 70.07 & 81.63 & 67.73 & 71.66 \\  \hline
GeoCLIP \citep{vivanco2024geoclip} 
    & RFF      &\underline{0.61} & \underline{0.84} & 0.37 & 0.78 & 70.56 & 81.65 & 67.78 & 71.93 \\
TaxaBind \citep{sastry2025taxabind}
    & RFF      & 0.60 & 0.77 & 0.38 & 0.72 & \textbf{72.15} & 80.58 & \textbf{68.18} & 71.71 \\  \hline
\model{} 
    & RFF      & \textbf{0.67} & \textbf{0.86} & \underline{0.40} & \textbf{0.82} & \underline{72.07} & \underline{81.76} & \underline{67.84} & \textbf{72.48}\\
\model{} \textit{Debias} 
    & RFF      & \textbf{0.67} & \underline{0.84} & \textbf{0.41} & \underline{0.81} & 72.03  & \textbf{81.88}   & \underline{67.84}  & \underline{72.17} \\    
\bottomrule
\end{tabular}}
\label{tab:loc_bench}
\end{table*}

\subsection{Location Benchmark Results}
We evaluate 2 location tasks. Table \ref{tab:loc_bench} shows all the experimental results.
\paragraph{Geo-Aware Image Regression.}  
We evaluate the effectiveness of location priors using the datasets from MOSAIKS \citep{rolf2021generalizable}, which includes four regression tasks: population density, forest cover, nightlight luminosity, and elevation estimation. The results indicate that incorporating location information significantly enhances model performance, yielding up to a 50\% improvement over models without location priors. Furthermore, using pretrained checkpoints further refines the learned geospatial representations. In particular, pretraining with \model{} provides the greatest gains, achieving improvements of 0.06, 0.02, 0.03, and 0.04 in $R^2$ for population density, forest cover, nightlight luminosity, and elevation regression, respectively.
\paragraph{Geo-Aware Species Recognition.}  
This task aims to classify images of different animal species. We use four datasets: BirdSnap \citep{berg2014birdsnap}, NABirds \citep{van2015building}, iNat2017 \citep{van2018inaturalist}, and iNat2018 \citep{van2018inaturalist}. Using a pretrained GeoCLIP encoder provides a moderate performance boost over random initialization. Notably, \model{} achieves the best results on NABirds and iNat2018 while remaining strongly competitive on BirdSnap and iNat2017 with TaxaBind. Noteably, \model{} is pretrained on our Streetscapes1M dataset that does not contain any species images but streetview images, whereas TaxaBind \citep{sastry2025taxabind} was pretrained on 2.55 million colocated satellite and ground-level species image pairs, giving TaxaBind a strong advantage for this task. These results clearly demonstrate the effectiveness of \model{}'s pretraining strategy and \textbf{strong out-of-domain generalizability}.  

\begin{table*}[t!]
\centering
\caption{Ablation study results on Street View and Remote Sensing benchmarks. For Street View tasks, non-linear probing RMSE $\downarrow$ is reported. For Remote Sensing tasks, tuning head only mIoU $\uparrow$ is reported.}
\label{tab:ablation_study_full}
\resizebox{\linewidth}{!}{%
\begin{tabular}{lcccc}
\toprule
\multirow{3}{*}{Ablation Settings} & \multicolumn{2}{c}{Street View Imagery Benchmark} & \multicolumn{2}{c}{Remote Sensing Benchmark} \\
\cmidrule(lr){2-3} \cmidrule(lr){4-5}
 & Socio-economic & Human Perception & Burn Scars & Crop Type \\
 & (RMSE $\downarrow$) & (RMSE $\downarrow$) & (mIoU $\uparrow$) & (mIoU $\uparrow$) \\
\midrule
\model{}-MAE                                   & 0.9011            & 2.0495           & 74.15             & 34.18 \\
\model{} w/o NILI and Loc \citep{huynh2025contrastive} & 0.8451    & 1.7056           & 80.45             & 52.78 \\
\model{} w/o NILI                               & 0.8588            & 1.7342           & 82.12             & 55.01 \\
\model{} w/o Loc                                & \underline{0.8352}& 1.6782      & \underline{82.94} & \underline{55.41} \\ \hline
\model{}-Bilinear                        & 0.8431     & \underline{1.6712}    & 82.05             & 55.12  \\
\model{}-Bicubic                         & 0.8435     & 1.6734    & 82.11             & 55.23  \\ \hline
\textbf{\model{}}                               & \textbf{0.8349}   & \textbf{1.6489}  & \textbf{83.26}    & \textbf{55.53} \\
\bottomrule
\end{tabular}
}
\end{table*}

\subsection{Ablation Studies}
To rigorously validate the contributions of each component in \model{}, we conduct a detailed ablation study (Table \ref{tab:ablation_study_full}) alongside our benchmark results (Tables \ref{tab:street_view_benchmark} and \ref{tab:remote_sensing_benchmark}). We can find three insights by analyzing these results.

First, explicit semantic alignment outperforms reconstruction. The \model{}-MAE variant, which treats street view images as masked patches for reconstruction, performs significantly worse than all contrastive variants. This confirms that pixel-level reconstruction is ill-suited for aligning geometrically disparate views, whereas our contrastive objective successfully captures high-level semantic correspondences. This performance gap is particularly obvious in the Burn Scars segmentation task, where the mIoU drops by over 9\% (from 83.26\% to 74.15\%) when using a reconstruction-based objective, highlighting the necessity of semantic-level supervision for specialized geospatial tasks.

Second, NILI is the primary driver of cross-scale alignment. Comparing the variants to the baseline method that lacks both implicit representation and location encoding (\model{} w/o NILI and Loc, akin to \citet{huynh2025contrastive}), we observe that simply adding the NILI module (\model{} w/o Loc) yields a substantial performance boost (e.g., mIoU improves from 52.78\% to 55.41\% on crop type classification task using RS images). Crucially, \model{} w/o Loc consistently outperforms \model{} w/o NILI across both tasks. This indicates that the visual alignment provided by NILI, which resolves the scale discrepancy between different views, is more critical for representation quality than global pooling strategies. Furthermore, to prove that this gain comes directly from our specific NILI design rather than just any spatial alignment, we test replacing NILI with standard deterministic interpolation methods (\model{}-Bilinear and \model{}-Bicubic). As shown in Table \ref{tab:ablation_study_full}, these naive interpolations yield very limited improvements over the \model{} w/o NILI baseline. For example, the mIoU only increases slightly from 55.01\% to 55.12\% (Bilinear) and 55.23\% (Bicubic) in crop type mapping task, and in burn scars estimation task the performance even drops from 82.12\% to 82.05\% (Bilinear) and 82.11\%(Bicubic). This clearly shows that traditional interpolation is too difficult to handle the perspective gap between ground-level and overhead views due to its discrete modeling nature, while NILI successfully learns fine-grained continuous spatial alignment. Similar trends are observed in socio-economic and human perception regression tasks, where NILI achieves a superior RMSE (0.8352/1.6782) compared to Bilinear interpolation (0.8431/1.6712) and Bicubic interpolation (0.8435/1.6734).

Finally, location encoding can further enhance the performance. While NILI handles the physical alignment, adding the location encoder completes the framework. The full \model{} achieves the best performance across all metrics. As shown in the Street View Benchmark (Table \ref{tab:street_view_benchmark}), the location encoder is particularly effective for abstract, spatially correlated tasks (e.g., human perception and socio-economic prediction), where geolocation serves as a strong prior to complement the visual features extracted by NILI. Quantitatively, integrating geolocation priors further reduces the RMSE for human perception from 1.6782 to 1.6489 and elevates the burn scars mIoU from 82.94\% to 83.26\%. Notably, \model{} avoids shortcut learning through coordinate memorization. While the location branch provides a beneficial geographic prior for socio-economic tasks, its removal (\model{} w/o Loc) still results in SOTA performance on pure visual tasks like remote sensing crop type classification. This confirms that our pre-training distills discriminative visual features rather than relying on geographic shortcuts.

\section{Discussions} 
\label{discussions}

\subsection{Multi-Modal Fusion}
\label{sec:multi-modal}
We investigate the impact of different modality combinations on model prediction accuracy for socio-economic indicator prediction tasks. Specifically, we compare two single-modality variants, i.e., using the street view (SV) image encoder alone, denoted as \model{}, and using the remote sensing (RS) image encoder alone, denoted as \model{} (RS Only), with several multimodal fusion strategies, including GAIR (RS+SV), GAIR (SV+Loc), and GAIR (SV+Loc+RS).

\begin{figure*}
  \centering
  \includegraphics[width=0.8\linewidth]{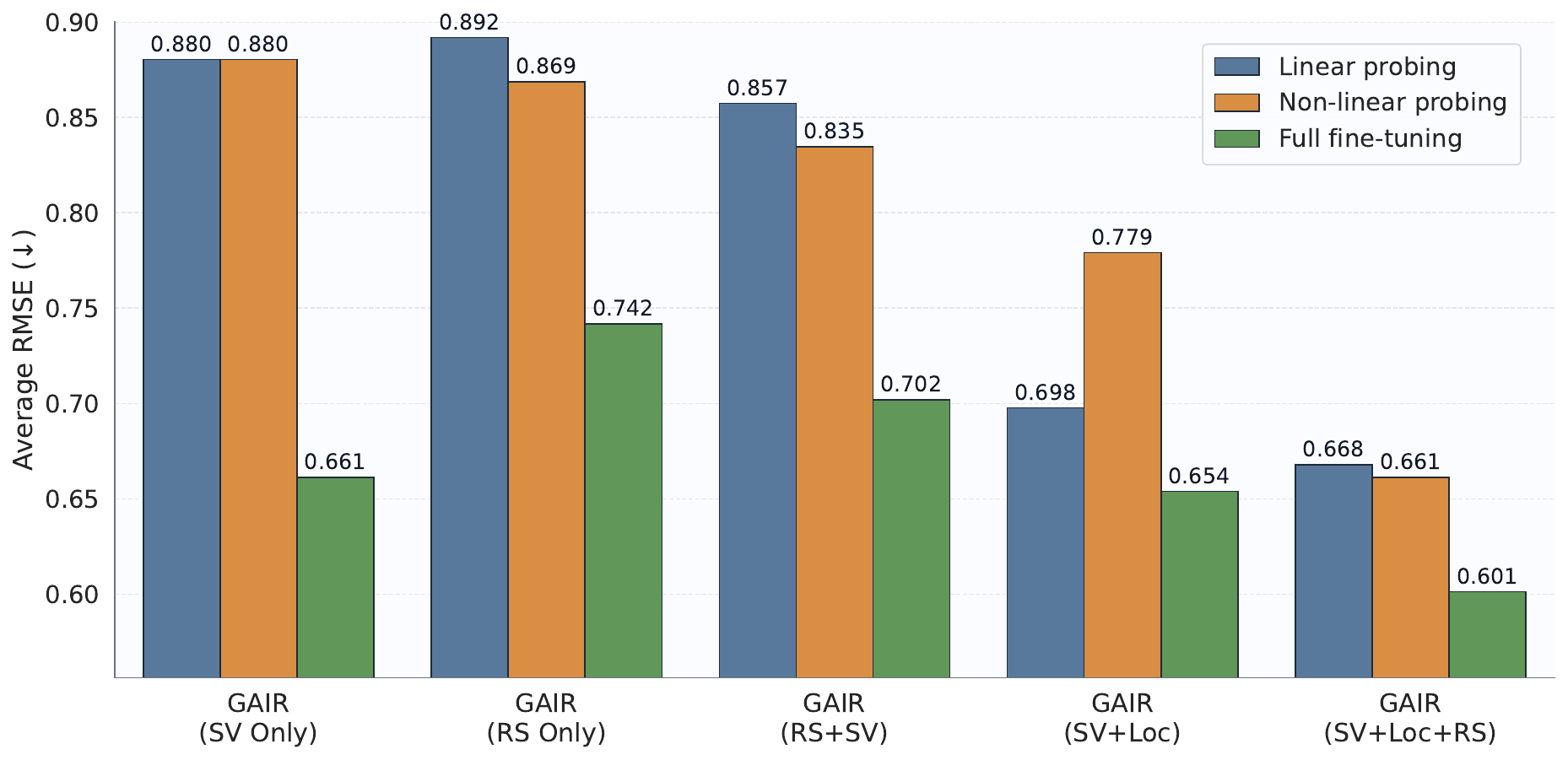}
  \caption{Average performance comparison of single-modality and multimodal fusion strategies over all socio-economic indicator prediction tasks.}
  \label{fig:multimodal_fusion}
\end{figure*}

GAIR (RS+SV) leverages the RS image encoder and the street view image encoder to extract geographically co-located RS and SV embeddings at location $x_i$, which are then concatenated and fed into a probing head for socio-economic indicator prediction. Similarly, GAIR (SV+Loc) combines embeddings from the SV image encoder and the location encoder, while GAIR (SV+Loc+RS) integrates all three modalities for prediction. The result is shown in Figure \ref{fig:multimodal_fusion}.

In particular, the full multimodal fusion model, GAIR (SV+Loc+RS), achieves the lowest average RMSE across all three evaluation settings: non-linear probing (0.6611), full fine-tuning (0.6012), and linear probing (0.6678). This corresponds to a substantial error reduction compared to both single-modality and dual-modality baselines. For example, relative to the single-modality baselines, \model{} (RS Only) and \model{} (SV Only), the full fusion model reduces the average RMSE by approximately 20\% across all settings. Moreover, even when compared with the strongest dual-modality baseline in each setting, GAIR (SV+Loc), the full fusion approach yields an additional average RMSE reduction ranging from 4\% to 15.1\%, highlighting the effectiveness of multimodal fusion.
\begin{figure}
    \centering
    \begin{subfigure}{0.49\linewidth}
        \includegraphics[width=\linewidth]{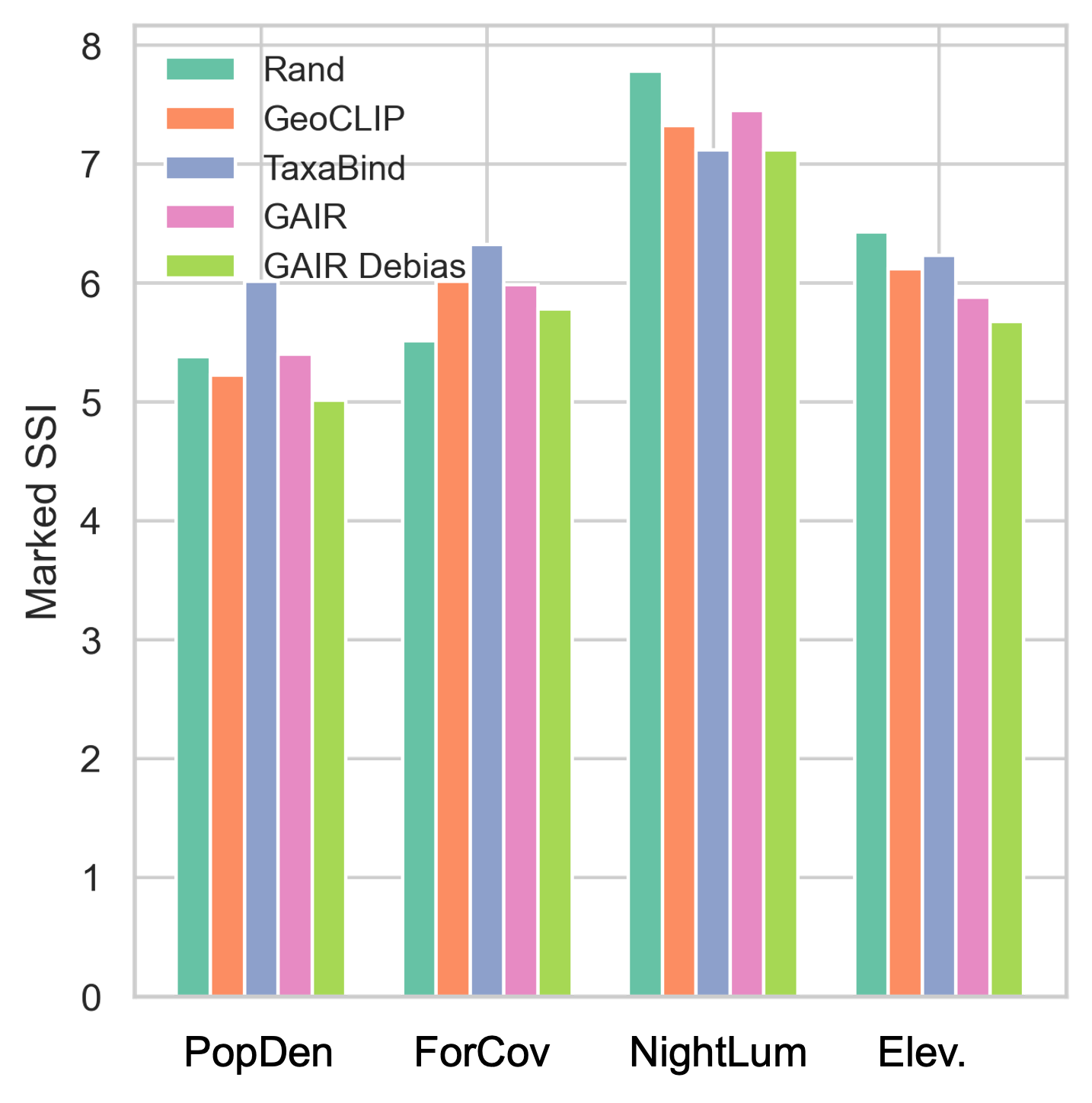}
        \caption{Image Regression}
        \label{fig:sub1}
    \end{subfigure}%
    \hspace{0.001\linewidth}
    \begin{subfigure}{0.49\linewidth}
        \includegraphics[width=\linewidth]{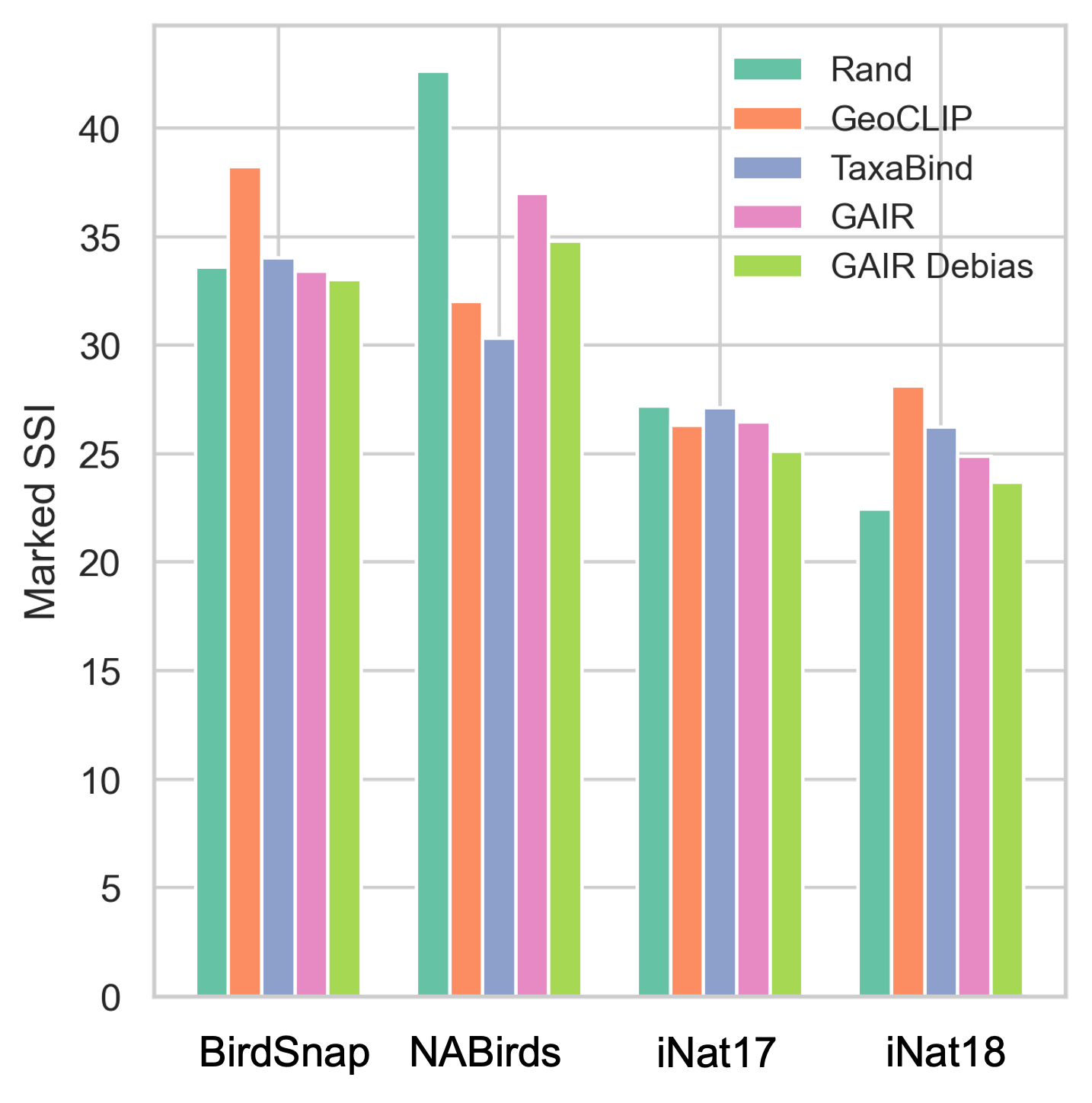}
        \caption{Species Recognition}
        \label{fig:sub2}
    \end{subfigure}
    \caption{Geo-Bias Score on the LocBench.}
    \label{fig:loc_bench_bias}
    \vspace{-18pt}
\end{figure}

\begin{figure*}[t]
    \centering
    \includegraphics[width=0.7\textwidth]{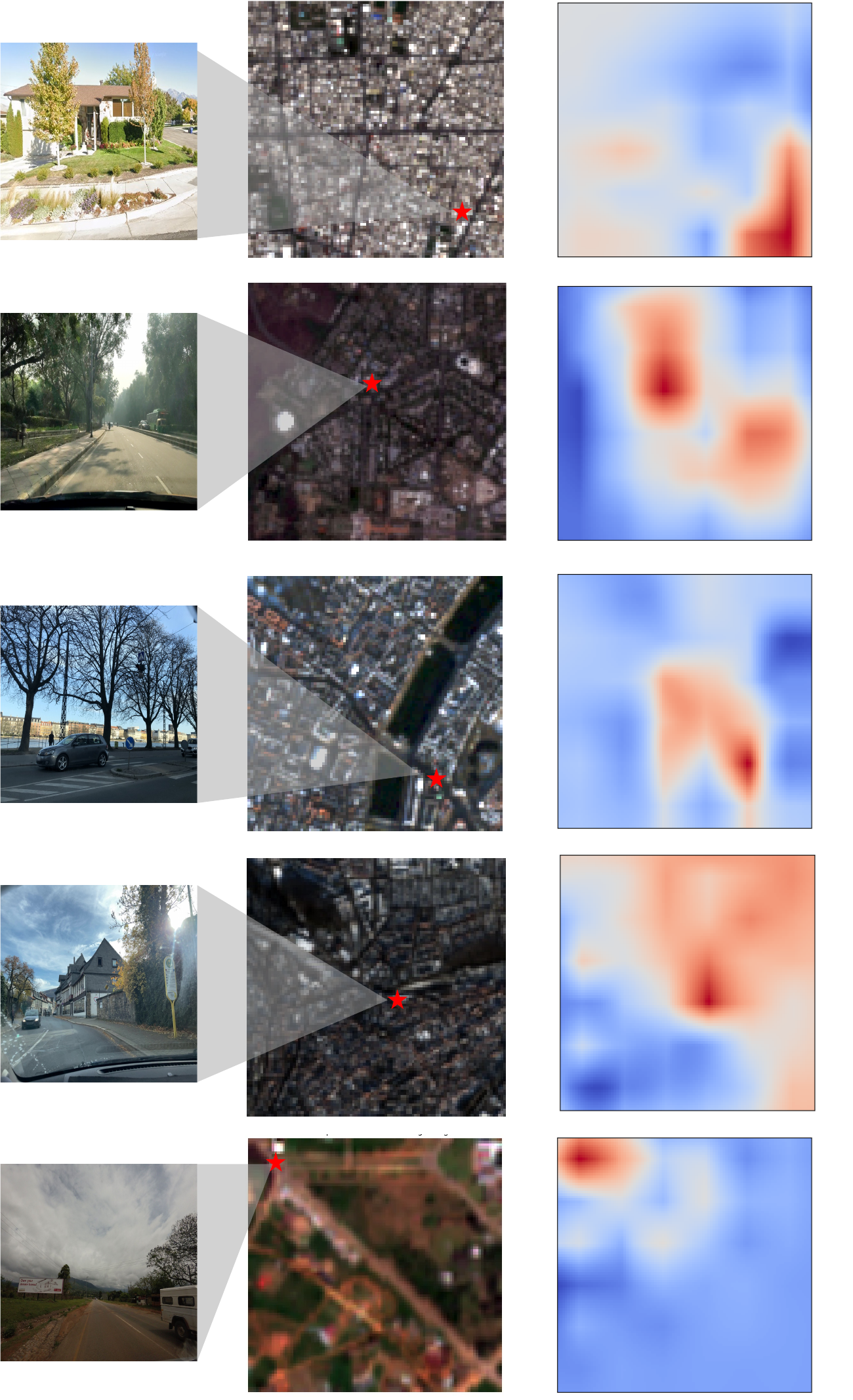}
    \caption{Visualization of cross-modal alignment in \model{} via heat maps. 
    Red stars mark the SV image location $x_i$. 
    Cosine similarity between $g(s_i)$ and localized RS embeddings $\hat{z}_i$.}
    \label{fig:spatial_alignment_rs_sv}
\end{figure*}
\begin{figure*}[t]
    \centering
    \includegraphics[width=1\textwidth]{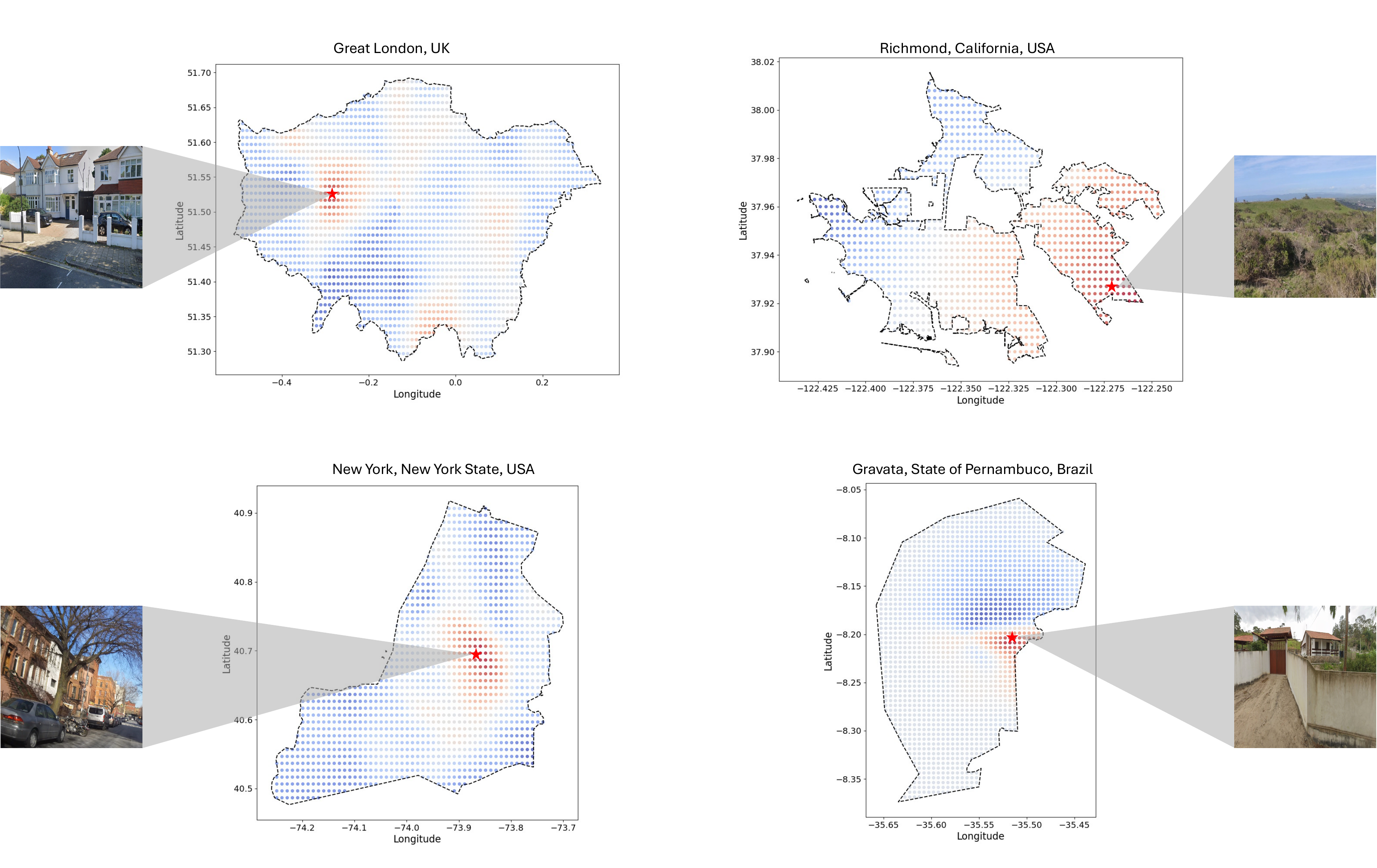}
    \caption{Visualization of cross-modal alignment in \model{} via heat maps. 
    Red stars mark the SV image location $x_i$. 
    Cosine similarity between $g(s_i)$ and location embeddings $e(x_i)$ sampled across Greater London.}
    \label{fig:spatial_alignment_sv_loc}
\end{figure*}

\subsection{Model Geographically Debiasing} 
Our Streetscapes1M dataset consists of triples containing street view images, their geolocations, and corresponding RS images. Due to the inherent constraints of street view data collection (e.g., road accessibility, commercial coverage), geographic coverage is significantly denser in urban areas while remaining sparse or non-existent in rural regions. This imbalance introduces a strong urban–rural bias. Consequently, pretraining \model{} exclusively on this dataset risks \textbf{amplifying geographic bias}, potentially degrading performance in underrepresented regions and limiting the model's global applicability \citep{wu2024torchspatial,manvi2024large}.

To mitigate this bias, \textbf{we use the factorized design of \model{}, which uniquely allows for the independent optimization of modality-specific encoders.} Unlike tightly coupled architectures that require strictly paired triplets, we address the rural data gap by collecting an additional 200,000 urban-rural balanced training samples consisting \textbf{solely} of remote sensing imagery and location metadata. We further pretrain the RS and location encoders on this balanced subset for 10 epochs, \textbf{excluding the SV encoder due to the unavailability of ground-level imagery in these regions.} \rone{More candidly, in extreme off-road or wilderness areas where street-view data is physically absent, the factorized architecture of \model{} naturally ``degenerates'' into pure remote-sensing or ``remote-sensing + location'' reasoning. Rather than being a weakness, this graceful degeneration reflects the flexibility of the architecture, i.e., the SV encoder is simply dropped while the remaining encoders continue operating in the same geo-aligned embedding space.}

The resulting model, denoted as \textbf{\model{} \emph{Debias}}, is compared against the original \model{} and baselines such as GeoCLIP \citep{vivanco2024geoclip} and TaxaBind \citep{sastry2025taxabind} on the remote sensing (Table \ref{tab:remote_sensing_benchmark}) and location benchmarks (Table \ref{tab:loc_bench}). We also evaluate geographic bias using the \textit{marked SSI score} (Geo-Bias Score) \cite{wu2024torchspatial} on location benchmark. Firstly, \textbf{we observe a significant reduction in geographic bias}, as shown in Figure \ref{fig:loc_bench_bias}. Lower marked SSI scores indicate reduced performance disparity across different geographic regions. \textbf{This demonstrates that leveraging easily accessible RS and location data can effectively mitigate geographic bias even in the absence of street view supervision, highlighting the flexibility of our framework.}

In location benchmark remote sensing benchmark, the performance of \model{} \emph{Debias} exhibits remarkable stability, remaining nearly unchanged compared to the original \model{}. Specifically, as shown in Table \ref{tab:remote_sensing_benchmark}, the debiased model preserves its state-of-the-art standing in RS tasks, even yielding slight improvements in linear probing (e.g., reaching 83.78\% mIoU on Burn Scars). Similarly, Table \ref{tab:loc_bench} confirms that across various regression (e.g., $R^2$ of 0.67 for PopDen) and classification tasks, the performance remains highly consistent. This stability is significant as it demonstrates that our debiasing strategy is effective and the \model{} \emph{Debias} is stable. 

\subsection{Analysis of Multimodal Geo-Alignment}

We examine whether \model{} effectively captures spatial relationships across different modalities. Specifically, we compute cosine similarities between a street view (SV) image embedding $g(s_i) \in \mathbb{R}^{D}$ and localized remote sensing (RS) embeddings $\rxi$ generated by the implicit neural representation module $\inrfun$ at different query locations $x_i^{(q)}$. As illustrated in Figure~\ref{fig:spatial_alignment_rs_sv}, \model{} learns to align SV and RS representations in a geographically consistent manner, with higher similarity concentrated around the true image location.

To further assess cross-modal alignment involving explicit geographic representations, we compare the SV embedding $g(s_i)$ with location embeddings $e(x_i)$ sampled on a uniform spatial mesh centered at the image location. As shown in Figure~\ref{fig:spatial_alignment_sv_loc}, the resulting similarity patterns are well aligned with street-level visual semantics, indicating that \model{} consistently associates SV features with their corresponding geographic context across modalities.

\subsection{Impact of Geolocation Noise on Geo-alignment}
\label{app:noise-importance}

A central idea of \model{} is that accurate geolocation provides the spatial anchor for aligning heterogeneous geospatial modalities. If the location metadata is corrupted, the alignment between SV and RS embeddings becomes unreliable, which should directly affect downstream task performance. To verify this, we introduce controlled noise into the GPS coordinates and measure the degradation.

\paragraph{Noise Injection.}
We use the von Mises–Fisher (vMF) distribution to simulate the location noise, because it is the standard choice for modeling random perturbations on a spherical surface. The concentration parameter $\kappa$ controls the noise level: a larger $\kappa$ means less GPS noise (a tighter vMF distribution), while a smaller $\kappa$ injects stronger positional errors. Then, starting from a pretrained checkpoint (100 epochs) of \model{}, we continue pretraining \model{} for 2 additional epochs with noisy coordinates, which are derived by adding a GPS noise to the ground truth locations drawn from vMF distributions with different $\kappa$ values. We then evaluate \model{} trained on noisy locations with different noise levels on the socio-economic indicator regression benchmark using non-linear probing.

\paragraph{Results and Discussions.}
Table~\ref{tab:geo-noise} shows that increasing location noise (corresponding to smaller $\kappa$ values) consistently leads to lower model performance. This trend highlights that geo-alignment is not optional but essential: when the model cannot align RS and SV features to the correct spatial anchor, its predictive power deteriorates.
The performance degradation under higher noise highlights the importance of precise geo-alignment: accurate coordinates enable consistent cross-modal fusion, whereas noisy locations disrupt this alignment. This experiment empirically supports our design choice of explicitly using geolocations to align different modalities in \model{}.

\begin{table}[t!]
\centering
\caption{Effect of geolocation noise on \model{} for the socio-economic regression tasks. A lower RMSE reflects better performance.}
\label{tab:geo-noise}
\small
\begin{tabular}{lcccc}
\toprule
\textbf{Noise Level} & No noise & $\kappa=300$ & $\kappa=100$ & $\kappa=1$ \\
\midrule
RMSE & 0.8349 & 0.8360 & 0.8488 & 0.8517 \\
\bottomrule
\end{tabular}
\end{table}

\subsection{Sensitivity Analysis of Remote Sensing Encoder}

To evaluate the robustness of \model{}, we conduct a sensitivity analysis on the remote sensing (RS) encoder by varying the patch size and input resolution under the standard pre-training protocol. As shown in Table \ref{tab:resolution_ablation}, \model{} exhibits distinct behaviors regarding these two factors:

\begin{itemize}
    \item \textbf{Robustness to Input Size:} Increasing the input resolution from $96 \times 96$ to $128 \times 128$ leads stable performance with small fluctuations. This indicates that the default $96 \times 96$ crop size already captures sufficient spatial context for robust feature extraction.
    
    \item \textbf{Sensitivity to Patch Size:} Conversely, increasing the patch size from $8 \times 8$ to $16 \times 16$ results in a significant performance drop. Because remote sensing imagery inherently contains complex and dense spatial information \citep{zheng2020foreground,zheng2023farseg++}, smaller patch sizes are critical to preserve the fine-grained, localized features required for dense prediction tasks.
\end{itemize}

\begin{table*}[h]
\caption{Ablation study on input resolutions and patch sizes for \model{}. All results are reported in mIoU $\uparrow$. The original configuration (Patch 8, $96 \times 96$) is compared against coarser patch size and higher input resolution.}
\label{tab:resolution_ablation}
\centering
\setlength{\tabcolsep}{5pt}
\resizebox{0.95\textwidth}{!}{%
\begin{tabular}{lc cccc cccc}
\toprule
\multirow{3}{*}{Model Config} & \multirow{3}{*}{Input Size} & \multicolumn{4}{c}{Fine-tuning Head} & \multicolumn{4}{c}{Full Fine-tuning} \\
\cmidrule(lr){3-6} \cmidrule(l){7-10} 
 &  & \multirow{2}{*}{Burn Scars} & \multirow{2}{*}{Crop. Poly.} & \multicolumn{2}{c}{Crop Type} & \multirow{2}{*}{Burn Scars} & \multirow{2}{*}{Crop. Poly.} & \multicolumn{2}{c}{Crop Type} \\
\cmidrule(lr){5-6} \cmidrule(l){9-10} 
 &  &  &  & Linear & L-TAE &  &  & Linear & L-TAE \\
\midrule
\model{} (P16) & $96 \times 96$   & 81.12 & 40.21 & 52.45 & 51.30 & 84.85 & 39.90 & 52.10 & 54.85 \\
\model{} (P8)  & $96 \times 96$   & \textbf{83.26} & \underline{43.35} & \textbf{55.53} & \underline{54.01} & \textbf{87.00} & \underline{42.51} & \underline{55.66} & \textbf{57.90} \\
\model{} (P8)  & $128 \times 128$ & \underline{83.15} & \textbf{43.58} & \underline{55.52} & \textbf{54.15} & \underline{86.24} & \textbf{42.92} & \textbf{56.05} & \underline{57.20} \\
\bottomrule
\end{tabular}%
}
\end{table*}

\subsection{Efficiency Analysis}
\label{app:compute-cost}

In this section, we will analyze the computation cost of the proposed NILI module.

It is important to note that the Neural Implicit Local Interpolation (NILI) module is primarily designed for the pretraining stage to learn geo-aligned representations. For standard downstream tasks (e.g., classification, segmentation), the NILI module is discarded, and the model functions simply as a standard ViT-B backbone. Consequently, as shown in Table \ref{tab:compute-nili}, our model incurs \textbf{identical} inference costs (GFLOPs and Latency) compared to other ViT-B based GeoFMs such as SatMAE \citep{cong2022satmae} and CROMA \citep{fuller2024croma}. 

An exception arises only if one opts for the joint multi-modal inference strategy (Section \ref{sec:multi-modal}), where NILI is re-enabled, introducing a minor latency increase (approx. 5ms).


\begin{table}[t]
\centering
\caption{Inference compute comparison. Since all methods utilize a ViT-B backbone, \model{} shares the same computational cost as baselines in the standard inference setting (without NILI). The "Joint w/ NILI" row represents the optional multi-modal setting (Section \ref{sec:multi-modal}).}
\label{tab:compute-nili}
\resizebox{1\columnwidth}{!}{
\begin{tabular}{p{4.8cm}ccc} 
\toprule
\textbf{Model} & \textbf{GFLOPs} & \textbf{Inference Setting} & \textbf{Latency (ms)} \\
\midrule
SatMAE, CROMA, \newline SpectralGPT, DOFA, etc. & 23.10 & Standard (ViT-B only) & 10.34 \\ \midrule
\model{} (Ours) & 23.10 & Standard (ViT-B only) & 10.34 \\
\model{} (Ours with Multimodal Inputs) & 23.12 & Optional: Joint w/ NILI & 16.01 \\
\bottomrule
\end{tabular}
}
\end{table}


\section{Conclusion}
\label{sec:conclusion}
In this paper, we introduce \model{}, a geo-aligned multimodal learning framework that integrates remote sensing imagery, street view imagery, and geolocation data to enhance geospatial representation learning. Our core innovation lies in the Neural Implicit Local Interpolation (NILI) module, which transcends traditional discrete interpolation by acting as a continuous cross-view semantic decoder. By explicitly resolving extreme scale discrepancies between remote sensing and street view modalities, NILI effectively bridges the profound viewpoint gap between overhead and ground-level observations. The robustness of this spatially explicit alignment strategy has been comprehensively demonstrated across 9 downstream tasks and 22 datasets. 

\rone{However, we also would like to highlight some limitations of \model{} which are worth further investigation. First, the upper bound of NILI's cross-view representation learning is inherently constrained by the degree of physical line-of-sight overlap and information visibility between the aerial and ground-level views. Semantics not co-observable from both viewpoints cannot be recovered regardless of decoder expressiveness. Moreover,} the current iteration of \model{} does not explicitly incorporate view direction (heading) information, and integrating such geometric constraints remains a promising direction for future work. \rone{The Streetscapes1M dataset, being crowdsourced, lacks panoramic (360°) or directional metadata, which prevents heading from being used as a conditioning signal. Once high-quality panoramic data becomes widely available, directly incorporating viewing angle or heading parameters as conditional inputs to the NILI module would be a crucial direction for further improving complex spatial reasoning and cross-view alignment accuracy.}

\section*{Declaration of Generative AI and AI-assisted technologies in the writing process}
During the preparation of this work, the author(s) used GPT-5 in order to improve readability and language. After using this tool/service, the author(s) reviewed and edited the content as needed and take(s) full responsibility for the content of the publication. 

\section*{Acknowledgement} 
This work is mainly funded by the National Science Foundation under Grant No. \# 2521631 -- A Statistics-Based Geographic Bias Quantification and Debiasing Framework for GeoAI and Foundation Models. Additional support is provided by NSF Grants 2125858 and the UT Good Systems Grand Challenge. 
Zeping Liu acknowledges the support of the Amazon AI PhD fellowship. 
Any opinions, findings, conclusions, or recommendations expressed in this material are those of the authors and do not necessarily reflect the views of the National Science Foundation.





\printcredits

\bibliographystyle{cas-model2-names}

\bibliography{cas-refs}

\end{document}